\documentclass[letterpaper]{article}
\usepackage{aaai2026}
\usepackage{times}
\usepackage{helvet}
\usepackage{courier}
\usepackage[hyphens]{url}
\usepackage[utf8]{inputenc}
\usepackage{graphicx}
\usepackage{natbib}
\usepackage{caption}
\usepackage{amsmath,amssymb,amsfonts}
\usepackage{newunicodechar}
\newunicodechar{✅}{\checkmark}

\usepackage{booktabs}
\usepackage{multirow}

\usepackage{algorithm}
\usepackage{algpseudocode}

\usepackage{xcolor}
\usepackage{float}

\usepackage{tikz}
\usepackage{pgfplots}
\pgfplotsset{compat=1.18} 
\usetikzlibrary{positioning,shapes.geometric,arrows.meta,backgrounds,fit,calc}

\usepackage[most]{tcolorbox}

\usepackage{newfloat}
\usepackage{listings}
\floatstyle{ruled}
\newfloat{listing}{tb}{lst}{}
\floatname{listing}{Listing}

\DeclareCaptionStyle{ruled}{labelfont=normalfont,labelsep=colon,strut=off}

\definecolor{primaryColor}{RGB}{0,87,184}
\definecolor{secondaryColor}{RGB}{255,140,0}
\definecolor{neutralColor}{gray}{0.6}

\graphicspath{{image/}}
\makeatletter
\IfFileExists{image/HiVA.png}%
  {\newcommand{\hiva}{\raisebox{-0.25\height}{\includegraphics[height=3ex]{HiVA}}}}%
  {\newcommand{\hiva}{}}
\makeatother

\nocopyright

\title{\hiva~HiVA: Self-organized Hierarchical Variable Agent via \\ Goal-driven Semantic-Topological Evolution}

\author{
    Jinzhou Tang\textsuperscript{\rm 1}\thanks{Equal contribution.}\quad
    Jusheng Zhang\textsuperscript{\rm 1}\footnotemark[1]\quad
    Qinhan Lv\textsuperscript{\rm 1}\footnotemark[1]\quad
    Sidi Liu\textsuperscript{\rm 1}\\
    Jing Yang\textsuperscript{\rm 1}\quad
    Chengpei Tang\textsuperscript{\rm 1}\quad
    Keze Wang\textsuperscript{\rm 1}\thanks{Corresponding author.}
}
\affiliations{
    \textsuperscript{\rm 1}Sun Yat-sen University\\
    kezewang@gmail.com
}

\begin{document}

\maketitle

\begin{abstract}
Autonomous agents play a crucial role in advancing Artificial General Intelligence, enabling problem decomposition and tool orchestration through Large Language Models (LLMs). However, existing paradigms face a critical trade-off. On one hand, reusable fixed workflows require manual reconfiguration upon environmental changes; on the other hand, flexible reactive loops fail to distill reasoning progress into transferable structures. We introduce Hierarchical Variable Agent (HiVA), a novel framework modeling agentic workflows as self-organized graphs with the Semantic-Topological Evolution (STEV) algorithm, which optimizes hybrid semantic-topological spaces using textual gradients as discrete-domain surrogates for backpropagation. The iterative process comprises Multi-Armed Bandit-infused forward routing, diagnostic gradient generation from environmental feedback, and coordinated updates that co-evolve individual semantics and topology for collective optimization in unknown environments. Experiments on dialogue, coding, Long-context Q\&A, mathematical, and agentic benchmarks demonstrate improvements of 5-10\% in task accuracy and enhanced resource efficiency over existing baselines, establishing HiVA's effectiveness in autonomous task execution.
\end{abstract}

\begin{links}
    \link{Code}{https://anonymous.4open.science/r/HiVA-60C6}
\end{links}

\section{Introduction}

\begin{figure*}[ht]
    \centering
    \includegraphics[width=\linewidth]{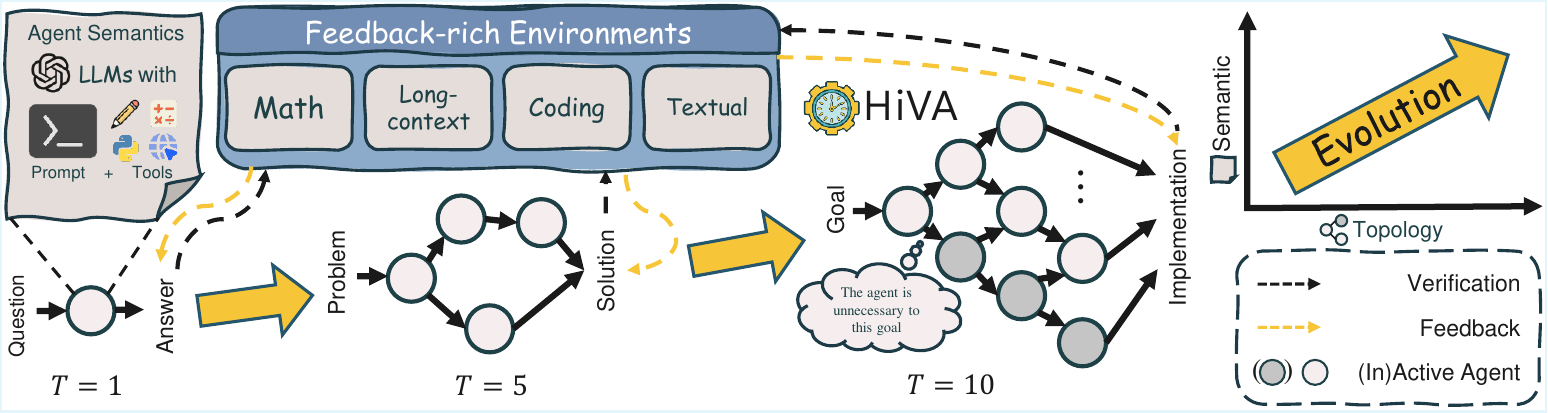}
    \caption{\textbf{Semantic-Topological Evolution from Singleton to Self-organized Complex Agents.} We explore how Large Language Models, when deployed in feedback-rich environments, can spontaneously form increasingly complex cognitive roles to refine their decision-making and tool use. Starting from a single agent with basic capabilities, HiVA fosters gradual evolution through semantic refinement and topological reconfiguration. Agents adapt to feedback-derived gradients from environmental interactions and develop into specialized yet interconnected sub-agents, thereby forming a complex system.}
    \label{fig:overview}
\end{figure*}

The pursuit of general-purpose autonomous agents, which are capable of independently solving complex, open-ended tasks, represents a central goal of artificial intelligence.~\cite{li2025ngentnextgenerationaiagents,bubeck2023sparksartificialgeneralintelligence,raman2025navigating} Large Language Models (LLMs) have emerged as a powerful backbone for such agents, enabling them to decompose goals, plan actions, and invoke tools through natural language reasoning.~\cite{li2024reviewprominentparadigmsllmbased} However, despite the success of LLM-based agents in applications such as automated software development and scientific discovery, their underlying multi-agent coordination paradigms remain fundamentally limited.~\cite{han2025llmmultiagentsystemschallenges,lamalfa2025largelanguagemodelsmiss}


Existing frameworks can be broadly categorized into two categories: (1) \textit{Manually-designed workflows}, which rely on fixed, structured agent interactions to complete tasks. While these offer modularity and reuse, they suffer from poor generalization to unseen task formats.~\cite{qin2022multiagent_transfer,hu2025owl}(2) \textit{Reactive agent loops} (\textit{e.g.}, ReAct~\cite{yao2022react}, AutoGPT~\cite{yang2023autogptonlinedecisionmaking}), which execute task solving as a sequence of language-based decisions. While more adaptable, these reactive agents fail to directly reuse reasoning patterns or continually improve themselves in dynamic environments~\cite{mialon2023augmented,DBLP:journals/corr/abs-2309-07864}. 
Though recent efforts have introduced mechanisms to optimize either agent behaviors or routing strategies~\cite{zhang2025agentic-supernet,zheng2025deepresearcherscalingdeepresearch}, they largely treat the coordination structure and semantics as independent. These approaches focus on incremental adaptation, i.e., tuning prompts, sampling structures, or adjusting routes from predefined templates, without a convincing mechanism to explore \textbf{how LLMs can spontaneously build a complex from a singleton}~\cite{guo2024large}. We argue that general-purpose agents should be \textbf{evolutionary} systems for dynamic environments: they must learn not only (1) \textit{what each agent should do} (i.e., semantic behavior), but also (2) \textit{how agents should interact and organize} (i.e., structural topology). This motivates our central question: \textit{Can a multi-agent system evolve both its internal semantics and collaborative structure} \textbf{from ZERO} \textit{to achieve scalable, adaptive, and self-organized intelligence across diverse tasks?}

To this end, we propose Hierarchical Variable Agent (HiVA), a novel multi-agent framework based on the \textit{Semantic-Topological Evolution (STEV)} algorithm. In HiVA, the coordination structure is modeled as a dynamic computational graph, and each agent is represented as a configurable LLM module. Optimization occurs in a hybrid space that consists of: (1) a \textit{semantic space} for learning agent-level behaviors (\textit{e.g.}, prompts, tool configurations)~\cite{yuksekgonul2025optimizing}, and (2) a \textit{topological space} that encodes which agents are connected and how information flows among them~\cite{DBLP:journals/corr/abs-2502-02533}. Each optimization round in HiVA involves three steps: (1) a \textit{Forward Pass}, where a task is routed through a dynamically constructed subgraph of agents; (2) a \textit{Textual Gradient Feedback}, where language-based diagnostics are generated based on the environmental feedback to approximate gradient-like signals for optimization in a non-differentiable space; and (3) a \textit{Coordinated Update}, where each agent adjusts both its internal semantic parameters and its structural links to other agents. This loop enables HiVA to progressively refine its capabilities and collaboration structure across tasks, resulting in a self-optimizing, specialized multi-agent system. Experimental results across diverse benchmarks show that HiVA generally outperforms static workflows, reactive loops, and mainstream multi-agent optimization algorithms, achieving better transferability and greater efficiency.

Our contributions are threefold: (i) We propose HiVA, the first framework to unify semantic and structural evolution from a singleton in LLM-based multi-agent systems; (ii) We introduce \textit{Semantic-Topological Evolution}, a general principle for co-evolving agent behavior and collaboration; and (iii) We design a dynamic routing and update mechanism based on bandit exploration and textual gradients. 

\section{Related Works}
\label{sec:relatedwork}

\paragraph{LLM-Based Multi-Agent Systems}
Multi-agent systems (MAS) powered by Large Language Models (LLMs) excel at complex tasks like software development and scientific discovery~\cite{he2024llmbasedmultiagentsystemssoftware,su2025headsbetteroneimproved,Jusheng1,Jusheng2,Jusheng3}. Recent works like ReAct~\cite{yao2022react} and AutoGPT~\cite{yang2023autogptonlinedecisionmaking} emphasize the reasoning and long-term planning capabilities of LLM-driven MAS, while lacking the potential for scalability through reproducing their chain-of-thought (CoT). MetaGPT~\cite{hong2023} uses shared message pools, and DyLAN~\cite{DBLP:journals/corr/abs-2310-02170} employs layered communication to boost efficiency. However, their static workflows limit adaptability to novel tasks. In contrast, our HiVA framework can dynamically evolve, thereby enhancing flexibility and scalability.

\paragraph{Dynamic Self-Improvement Mechanisms}
Dynamic self-improvement mechanisms enable agents to adapt through feedback. Existing approaches can be categorized into two paradigms: the first involves text-based optimization methods, including both single-agent semantic~\cite{zhang2023} and multi-agent collaborative optimization ~\cite{wang2023}; the second employs reinforcement learning approaches for agent improvement~\cite{guo2025deepseek}. In contrast, HiVA enables co-evolution of agent parameters and topology from a singleton, achieving effective test-time-scaling for diverse tasks and scenarios.

\paragraph{Hierarchies and Topological Optimization}
Hierarchical structures and topological optimization enhance MAS efficiency. MetaGPT~\cite{hong2023} and MASAI~\cite{masai2024} streamline workflows with modular designs. MASS~\cite{guo2025} proposes a Multi-Agent System Search for prompt and topology optimization. G-designer~\cite{gdesigner2024} uses graph neural networks for topology optimization without semantic integration. HiVA’s Bayesian-driven hierarchical evolution enables better modeling of individuals in self-organized MAS.

\begin{figure*}[ht]
    \centering
    \includegraphics[width=\linewidth]{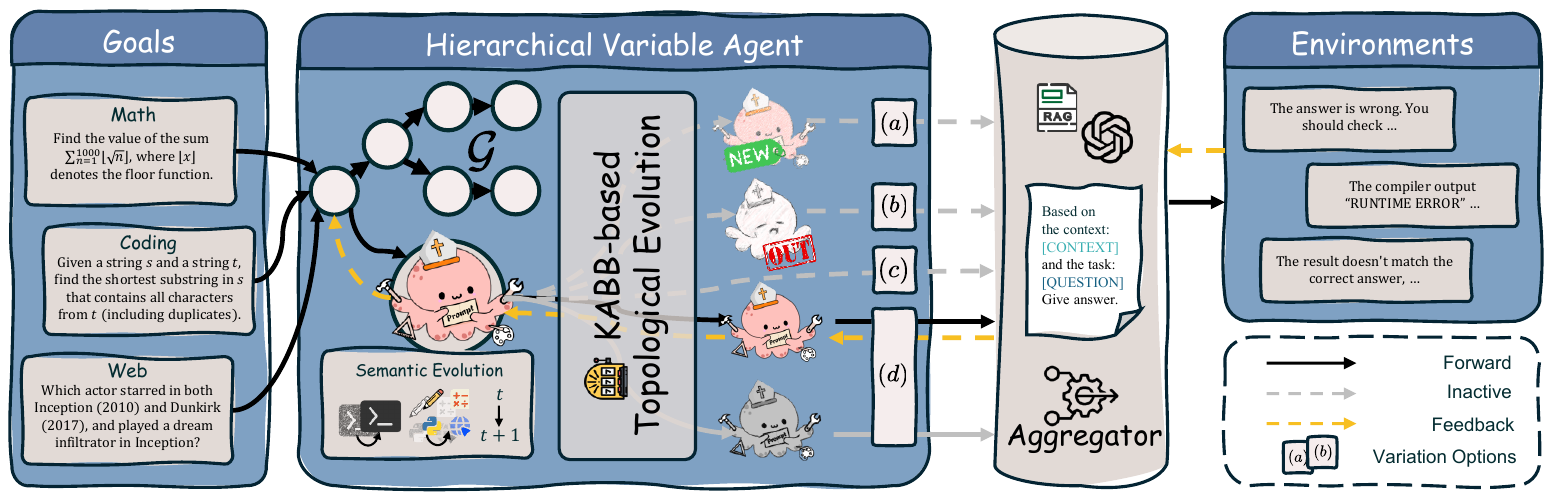}
    \caption{\textbf{Evolving mechanism of Hierarchical Variable Agent in a single iteration.} Driven by the goal, HiVA decomposes evolution into two stages: forward and backward propagation. In the forward pass, each agent in $\mathcal{G}$ utilizes KABB to select relevant successors and generate instructions for them. The aggregator then uses specific tools (\textit{e.g.}, RAG) to access MAS context and generate final answers. In the backward pass, each agent receives feedback from its successor (notably, aggregator receives feedback from the environments) and evolves their semantics (prompts and tools) and topologies with four options: (a) add a successor, (b) delete a successor, (c) connect directly to the aggregator, and (d) do nothing.}
    \label{fig:method}
\end{figure*}

\section{Methodology}

This paper introduces HiVA (Hierarchical Variable Agent), a novel framework designed to address the problem of adaptive task-flow optimization in Multi-Agent Systems (MAS). The core idea is to formulate this optimization problem as a process of generalized gradient descent within a \textbf{Hybrid Space}, which is constituted by both a semantic space and a topological space. The theoretical cornerstone of our work is the \textbf{Semantic-Topological Evolution (STEV)} algorithm, as illustrated in Figure \ref{fig:overview}, which enables optimization in the discrete and non-differentiable space of agentic graphs. Our STEV algorithm facilitates the simultaneous co-evolution of agents' internal parameters (semantics) and their collaborative structures (topology) by introducing \textbf{Textual Gradients} as a substitute for traditional gradients, combined with knowledge-aware dynamic computation graph construction. 

\subsection{Semantic-Topological TextGrad}

We define a complete solution for a multi-agent system as a point $s$ within a hybrid space $\mathcal{S} = \mathcal{G} \times \mathcal{P}_\Theta$, where $\mathcal{G}$ is the space of graph topologies and $\mathcal{P}_\Theta$ is the space of semantic parameters $\Theta$. The optimization objective is to find a solution $s^*$ that minimizes a black-box objective function $\mathcal{L}(s)$.
\[
s^* = \arg\min_{s \in \mathcal{S}} \mathcal{L}(s)
\]
This optimization is non-trivial, as the discrete and non-Euclidean nature of $\mathcal{S}$ renders the traditional gradient $\nabla_s \mathcal{L}$ ill-defined. To navigate this, we introduce the \textbf{Textual Gradient}~\cite{yuksekgonul2025optimizing} as a functional substitute. We operationalize it using a \textbf{Textual Gradient Parser}, implemented via an LLM, which translates raw textual feedback into a structured update instruction, $\Delta s_t$. The parser diagnoses whether the gradient applies to an agent's semantics or its topology, generating a corresponding command for either the semantic evolution function ($f_P$) or the topological evolution function ($f_G$). This reframes the optimization as a \textbf{Generalized Gradient Descent} process, with an update rule $s_{t+1} \leftarrow s_t \oplus \Delta s_t$, where the parsed gradient guides targeted updates across the hybrid space.

\subsection{The Environment: Oracle and Adversary}

The interaction space for our optimization problem is defined by the \textbf{Environment}, $\mathcal{E}_{\text{env}}$. Formally, the environment is a function that accepts the system's final output, $S_{\text{out}} = s(I_{\text{task}})$, and returns a potentially complex outcome or feedback. The environment plays a crucial dual role.

First, it acts as an \textbf{oracle}, providing the ground-truth feedback necessary for learning. The textual loss required for optimization is computed by an objective function $\mathcal{L}$ that maps the environment's rich dynamics driven by $s$ to a explainable signal: $\mathcal{L}(s) = \mathcal{L}(\mathcal{E}_{\text{env}}(s(I_{\text{task}})))$. 

Second, the environment acts as an \textbf{adversary}, defining the complexity and challenge of the task space. The robustness and adaptability of a solution $s$ are measured by its ability to consistently achieve low loss across the distribution of problems posed by $\mathcal{E}_{\text{env}}$. The nature of $\mathcal{E}_{\text{env}}$ can vary widely, encompassing: (a) programmatic environments with verifiable outcomes (\textit{e.g.}, code compilation, unit test execution); (b) data-driven environments where feedback is derived from metrics on a held-out dataset (\textit{e.g.}, QA); (c) agentic environments that provide interactive simulators or interfaces (\textit{e.g.}, browser, game, IDE); and (d) mathematical environments where feedback is qualitative text from a verifiable math evaluator (\textit{e.g.}, Lean). A successful optimization must yield a solution robust to the specific challenges of its target environment.

\subsection{The Iterative Optimization of HiVA}

The optimization process in HiVA is an iterative loop driving the solution $s_t = (G_t, \Theta_t)$ toward a minimum of the objective function, as illustrated in Algorithm \ref{algo:stev}. Each iteration consists of a coupled forward and backward pass. The function $\text{RepairTopology}(\cdot)$ removes isolated nodes and prevents cycles; more details can be found in the Algorithmic Details section of the Appendix.

\begin{algorithm}
\caption{\textsc{Semantic-Topological Evolution}}
\label{algo:stev}
\begin{algorithmic}
\State \textbf{Definitions}:
\State \quad \textit{Agent} \( v_i \in \mathcal{V} \): Entity with prompt \( p_i \), tool \( \tau_i \), mapping instruction \( x_i \) to output \( y_i \).
\State \quad \textit{Aggregator} \( v_a \): Aggregates outputs.
\State \quad \textit{Network} \(\mathcal{G}\): DAG with source \( v_s \) and sink \( v_a \).

\State \textbf{Input}: Instruction \( x_0 \), environment function \( \mathcal{E}_{\text{env}} \).
\State \textbf{Output}: Optimized \(\mathcal{G}\), result \( y \).

\State \textbf{Initialize}: Create \( v_s \) ( \( p_s \), \( t_s \)), connect \( v_s \to v_a \).
\For{$t = 1$ to $T$}
    \State \textbf{\ \ Forward}: Execute \textsc{ForwardPass}($\mathcal{G}, x_0$)
    \State \textbf{Loss}: \( \mathcal{L}_t \gets \mathcal{L}(\mathcal{E}_{\text{env}}(y)) \), \( \nabla_{\text{text}}\mathcal{L}_t \gets \text{LLM}(y, \mathcal{L}_t) \)
    \State \textbf{Backward}:
    \For{each \( v_i \in \text{reverse}(\sigma) \)}
        \State \( \frac{\partial\mathcal{L}_t}{\partial v_i} \gets \nabla_{\text{text}}\mathcal{L}_t \) if \( v_i = v_a \), else 
        \State \( \frac{\partial\mathcal{L}_t}{\partial v_i} \gets \text{LLM}(\{\frac{\partial\mathcal{L}_t}{\partial v_j} \mid v_j \in \text{successors}(v_i)\}, y_i) \)
        \State \( p_i, \tau_i \gets f_P(p_i, \tau_i, \frac{\partial\mathcal{L}_t}{\partial v_i}) \)
        \State \( \mathcal{G} \gets f_G(\mathcal{G},\{\mathcal{R}_{ij}\}_{j\in \text{successor}(v_i)}, \frac{\partial\mathcal{L}_t}{\partial v_i}) \) 
    \EndFor
    \State \( \alpha_i, \beta_i \leftarrow \alpha_i^{(t+1)}, \beta_i^{(t+1)} \)
    \State \( \mathcal{G} \gets \text{RepairTopology}(\mathcal{G}) \)
\EndFor
\State \textbf{Return}: \(\mathcal{G}, y\)
\end{algorithmic}
\end{algorithm}

The \textbf{Forward Pass}, as shown in Algorithm \ref{alg:forward}, probes the potential of the current solution $s_t$. It employs a dynamic routing mechanism to construct a task-specific execution subgraph $\mathcal{G}_{\text{exec},t}$, culminating in the generation of a final output $S_{\text{out},t}$.

This output is then subjected to the \textbf{Loss Evaluation} phase. It is passed to the environment, $\mathcal{E}_{\text{env}}$, and the resulting outcome is mapped to a textual loss: 
$$\mathcal{L}_t = \mathcal{L}(\mathcal{E}_{\text{env}}(S_{\text{out},t})).$$
The loss $\mathcal{L}_t$ triggers the \textbf{Backward Pass}. The process begins by estimating a global textual gradient at the sink point $v_a$ of $\mathcal{G}$ based on the environmental feedback. This gradient is then decomposed and propagated via a textual chain rule to each participating node, yielding localized textual gradients.

This backward flow of information culminates in a \textbf{Coordinated Update} of the current solution $s_t$. The semantic parameters and graph topology are co-optimized via textual gradient descent. Finally, the dynamic routing policy is refined by updating its Bayesian belief parameters, following KABB~\cite{zhang2025kabb}. 

\subsection{Knowledge-Based Subgraph Generation}

Dynamic routing underpins HiVA’s efficiency and adaptability in selecting agents for task-specific execution subgraphs. It models agent selection as a Multi-Armed Bandit (MAB) problem, solved using Thompson Sampling. With topological order, at each decision point, the system assigns $P_i$ to each agent $A_i$ by sampling from a probability distribution proportional to:
\[
P_i \propto \frac{\alpha_i^{(t)}}{\alpha_i^{(t)} + \beta_i^{(t)}} \cdot \exp\left(-\lambda \cdot \text{Dist}(A_i, I_{\text{task}})\right) \cdot \zeta(\mathcal{S}_t)^\eta
\]
Here, $\zeta(\mathcal{S}_t) = \frac{1}{|\mathcal{S}_t|(|\mathcal{S}_t|-1)} \sum_{v_i, v_j \in \mathcal{S}_t, i\neq j} C_{\text{syn}}^{(t)}(v_i, v_j)$ quantifies the collaborative effect within the selected subset $\mathcal{S}_t$ which include previous selected agents and $A_i$, where $C_{\text{syn}}^{(t)}(v_i, v_j)$ is the synergy gain coefficient for pairwise agent interactions, and $\eta$ adjusts its influence. This distribution balances historical performance (via Bayesian belief parameters $\alpha_i, \beta_i$), task relevance, and team synergy, penalized by a \textbf{knowledge-based cost function}:
\[
\text{Dist}(A_i, I_{\text{task}}) = \log(1 + d_I) \cdot \sum_{k=1}^4 \omega_k \Psi_k
\]
This cost function uses an external knowledge graph to measure the mismatch $\Psi_k$ across four sub-indicators of an agent’s capabilities against task requirements. The routing mechanism then constructs efficient, task-specific execution subgraphs $G_{\text{exec},t}$ through Thompson Sampling, which serve as the foundation for forward propagation. More details can be found in the Appendix on Knowledge-Based Cost Function.

The routing policy evolves by updating the Bayesian belief parameters for each agent based on performance and task alignment:
\[
\alpha_i^{(t+1)} = \gamma^{\Delta t} \alpha_i^{(t)} + \left[ r_i^{(t)} + \delta \cdot \text{KM}(A_i, I_{\text{task}}) \right] \cdot \mathbb{I}_{\{A_i \in \mathcal{S}_t\}}
\]
\[
\beta_i^{(t+1)} = \gamma^{\Delta t} \beta_i^{(t)} + \left[ 1 - r_i^{(t)} + \delta \cdot \text{KD}(A_i, I_{\text{task}}) \right] \cdot \mathbb{I}_{\{A_i \in \mathcal{S}_t\}}
\]
Here, $\text{KM}(A_i, I_{\text{task}}) = \rho_{\text{overlap}} \cdot \zeta(\{A_i, I_{\text{task}}\})$ measures task alignment and $\text{KD}(A_i, I_{\text{task}}) = 1 - \text{KM}(A_i, I_{\text{task}})$ plays the opposite, incorporating a synergy term scaled by a task relevance factor $\rho_{\text{overlap}}$. The updates combine a reward signal $r_i^{(t)}$ reflecting the agent’s contribution, a knowledge-driven adjustment $\text{KM}$, and an exponential decay factor $\gamma^{\Delta t} = e^{-\kappa \Delta t}$ to prioritize recent performance. The indicator function $\mathbb{I}_{\{A_i \in \mathcal{S}_t\}}$ ensures only agents in the selected subset are updated. This continuous learning process refines the policy, favoring agents that perform well and align closely with the task while promoting collaborative subsets, leading to increasingly effective subgraph generation over time.

\subsection{Multi-agent Structure as Memory}

In our HiVA framework, the multi-agent structure serves not only as an organizational scheme for computational units but also functions as the core memory mechanism of the system. Unlike traditional multi-agent systems that confine memory to internal states of individual agents, HiVA encodes collective memory within the network topology $\mathcal{G}$ and inter-agent connection weights, achieving distributed memory storage and retrieval.

\begin{algorithm}
\caption{\textsc{ForwardPass}}
\label{alg:forward}
\begin{algorithmic}
\State Initialize \( \mathcal{R}_{ij} \gets 0, \forall v_i,v_j \in \mathcal{V} \)
\State Compute topological order \(\sigma\) of \(\mathcal{G}\)
\For{each \( v_i \in \sigma \)}
    \If{\( v_i \) receives no instruction}
        \State \textbf{continue}
    \EndIf
    \State \( y_i \gets \tau_i(x_i) \)
    \State \( s_j \sim P_i \), \(v_j \in \text{successor}(v_i)\)
    \State \( \mathcal{V}' = \text{top-}k_{v_j\in \text{successor}(v_i)}(s_j) \)
    \For{each \( v_j \in \mathcal{V}' \)}
        \State \( \mathcal{R}_{ij} \gets \mathcal{R}_{ij} + 1 \)
        \State Generate \( x_j \gets \text{LLM}(x_i, y_i, p_j) \)
    \EndFor
\EndFor
\State \( y \gets v_a.\text{aggregate}(\{y_i \mid \text{deg}^{+}(v_i) = 1, \mathcal{R}_{ia}=1\}) \)
\State \textbf{Return}: \( y \)
\end{algorithmic}
\end{algorithm}

Specifically, each edge $(v_i, v_j) \in \mathcal{E}$ in the network topology $\mathcal{G} = (\mathcal{V}, \mathcal{E})$ carries a weight $C_{\text{syn}}^{(t+1)}(v_i, v_j)$ that encodes the historical accuracy and task relevance of information transfer from agent $v_i$ to $v_j$. These weights evolve continuously through a Bayesian update mechanism:
\[
C_{\text{syn}}^{(t+1)}(v_i, v_j) = C_{\text{syn}}^{(t)}(v_i, v_j) + \gamma \cdot \frac{\alpha_{ij}^{(t)}}{\alpha_{ij}^{(t)} + \beta_{ij}^{(t)}} \cdot \mathcal{R}_{ij}^{(t)}
\]
where $\mathcal{R}_{ij}^{(t)}$ represents the task contribution of edge $(v_i, v_j)$ in the $t$-th iteration, and $\gamma$ is the learning rate.

Notably, HiVA's memory mechanism exhibits hierarchical characteristics. At the macro level, the topological structure $\mathcal{G}$ stores long-term memory of collaboration patterns among agents; at the meso level, edge weights $w_{ij}$ record the effectiveness of specific collaboration paths; at the micro level, agents' semantic parameters $\Theta_i$ preserve individual specialized knowledge. This ensures that the system can utilize historical experience at different granularities.

\subsection{Semantic-Topological Evolution}

HiVA's core innovation lies in simultaneously optimizing agent semantic parameters and network topology from singleton through two complementary functions $f_P$ and $f_G$, as illustrated in Figure~\ref{fig:method}. Detailed prompting strategies can be found in the Prompting Strategies section of the Appendix.

\textbf{Semantic evolution function} $f_P$ handles semantic parameter evolution. Given textual gradient $\frac{\partial\mathcal{L}_t}{\partial v_i}$, current semantic parameters $p_t$, and tool configuration $t_i$, it produces updated parameters. The LLM analyzes the feedback to identify potential improvements in the agent's prompt and tool definition, and then generates refined prompts that address issues.

\textbf{Topological evolution function} $f_G$ modifies the local topology at successor nodes of $v_i$ based on textual gradients, topological connections, and task contribution matrix $\mathcal{R}$. Further, $f_G$ instructs an LLM to analyze local topological neighborhoods and determine optimal structural modifications, including adding connections, removing redundant edges, or restructuring subgraphs.

\begin{table*}[t]
\caption{Performance comparison across different task-driven environments on \texttt{Qwen-2.5-72B-Instruct-Turbo}. Results show accuracy scores (\%) for each method across programmatic, textual, long-context, and mathematical reasoning tasks. Bold indicates best performance. Subscript values denote standard deviation across three runs.}
\label{tab:task_performance}
\resizebox{\textwidth}{!}{%
\begin{tabular}{@{}l*{8}{c}c@{}}
\toprule
\multirow{2}{*}{\textbf{Method}} & \multicolumn{2}{c}{\textbf{Mathematical}} & \multicolumn{2}{c}{\textbf{Long-context}} & \multicolumn{2}{c}{\textbf{Programmatic}} & \multicolumn{2}{c}{\textbf{Textual}} & \multirow{2}{*}{\textbf{Avg.}} \\
\cmidrule(lr){2-3} \cmidrule(lr){4-5} \cmidrule(lr){6-7} \cmidrule(lr){8-9}
& MATH & GSM-8K & HotpotQA & 2WikihopQA & HumanEval & MBPP & MMLU & BBH & \\
\midrule
Vanilla & 82.7 & 91.2 & 67.4 & 76.2 & 86.1 & 86.7 & 86.1 & 86.3 & 82.6 \\
\midrule
CoT & 84.3\textsubscript{\tiny $\uparrow$1.6\%} & 92.1\textsubscript{\tiny $\uparrow$0.9\%} & 68.8\textsubscript{\tiny $\uparrow$1.9\%} & 77.6\textsubscript{\tiny $\uparrow$1.7\%} & 87.4\textsubscript{\tiny $\uparrow$1.5\%} & 87.3\textsubscript{\tiny $\uparrow$0.9\%} & 85.2\textsubscript{\tiny $\downarrow$0.7\%} & 87.6\textsubscript{\tiny $\uparrow$1.4\%} & 83.8\textsubscript{\tiny $\uparrow$1.5\%} \\
Self-Consistency & 84.7\textsubscript{\tiny $\uparrow$2.2\%} & 92.6\textsubscript{\tiny $\uparrow$1.4\%} & 69.1\textsubscript{\tiny $\uparrow$2.4\%} & 78.3\textsubscript{\tiny $\uparrow$2.4\%} & 88.2\textsubscript{\tiny $\uparrow$2.4\%} & 88.2\textsubscript{\tiny $\uparrow$1.5\%} & 85.6\textsubscript{\tiny $\downarrow$0.3\%} & 88.1\textsubscript{\tiny $\uparrow$2.0\%} & 84.4\textsubscript{\tiny $\uparrow$2.2\%} \\
Self-Refine & 85.1\textsubscript{\tiny $\uparrow$2.8\%} & 93.2\textsubscript{\tiny $\uparrow$2.0\%} & 69.4\textsubscript{\tiny $\uparrow$3.1\%} & 78.6\textsubscript{\tiny $\uparrow$3.0\%} & 86.2\textsubscript{\tiny $\uparrow$0.1\%} & 87.1\textsubscript{\tiny $\uparrow$0.3\%} & 85.1\textsubscript{\tiny $\downarrow$1.3\%} & 87.2\textsubscript{\tiny $\uparrow$0.8\%} & 84.0\textsubscript{\tiny $\uparrow$1.7\%} \\
Multi-Agent Debate & 85.4\textsubscript{\tiny $\uparrow$3.1\%} & 93.3\textsubscript{\tiny $\uparrow$2.2\%} & 70.2\textsubscript{\tiny $\uparrow$3.9\%} & 79.1\textsubscript{\tiny $\uparrow$3.7\%} & 88.1\textsubscript{\tiny $\uparrow$2.3\%} & 87.4\textsubscript{\tiny $\uparrow$0.9\%} & 85.4\textsubscript{\tiny $\downarrow$0.7\%} & 87.6\textsubscript{\tiny $\uparrow$1.4\%} & 84.6\textsubscript{\tiny $\uparrow$2.4\%} \\
\midrule
DyLAN & 85.3\textsubscript{\tiny $\uparrow$3.0\%} & 92.9\textsubscript{\tiny $\uparrow$1.8\%} & 69.7\textsubscript{\tiny $\uparrow$3.1\%} & 78.8\textsubscript{\tiny $\uparrow$3.3\%} & 89.7\textsubscript{\tiny $\uparrow$4.2\%} & 87.3\textsubscript{\tiny $\uparrow$0.6\%} & 85.2\textsubscript{\tiny $\downarrow$1.3\%} & 87.1\textsubscript{\tiny $\uparrow$0.8\%} & 84.5\textsubscript{\tiny $\uparrow$2.3\%} \\
AgentVerse & 85.6\textsubscript{\tiny $\uparrow$3.3\%} & 93.1\textsubscript{\tiny $\uparrow$2.1\%} & 70.1\textsubscript{\tiny $\uparrow$3.9\%} & 79.3\textsubscript{\tiny $\uparrow$3.9\%} & 89.6\textsubscript{\tiny $\uparrow$4.1\%} & 87.6\textsubscript{\tiny $\uparrow$0.9\%} & 85.7\textsubscript{\tiny $\downarrow$0.7\%} & 87.4\textsubscript{\tiny $\uparrow$1.4\%} & 84.8\textsubscript{\tiny $\uparrow$2.7\%} \\
ADAS & \textbf{86.1\textsubscript{\tiny $\uparrow$4.0\%}} & 93.4\textsubscript{\tiny $\uparrow$2.5\%} & 72.3\textsubscript{\tiny $\uparrow$6.8\%} & 80.7\textsubscript{\tiny $\uparrow$5.6\%} & 85.2\textsubscript{\tiny $\downarrow$1.0\%} & 89.2\textsubscript{\tiny $\uparrow$2.8\%} & 87.2\textsubscript{\tiny $\uparrow$1.0\%} & 88.6\textsubscript{\tiny $\uparrow$2.5\%} & 85.3\textsubscript{\tiny $\uparrow$3.3\%} \\
MaAS & 85.7\textsubscript{\tiny $\uparrow$3.6\%} & 94.1\textsubscript{\tiny $\uparrow$3.2\%} & 76.2\textsubscript{\tiny $\uparrow$13.1\%} & 81.1\textsubscript{\tiny $\uparrow$6.4\%} & 92.3\textsubscript{\tiny $\uparrow$7.2\%} & 90.1\textsubscript{\tiny $\uparrow$3.9\%} & 89.4\textsubscript{\tiny $\uparrow$3.8\%} & 90.6\textsubscript{\tiny $\uparrow$5.0\%} & 87.4\textsubscript{\tiny $\uparrow$5.8\%} \\
\midrule
\textbf{HiVA (ours)} & 81.2\textsubscript{\tiny $\downarrow$1.8\%} & \textbf{94.5\textsubscript{\tiny $\uparrow$3.6\%}} & \textbf{79.7\textsubscript{\tiny $\uparrow$18.3\%}} & \textbf{86.5\textsubscript{\tiny $\uparrow$13.5\%}} & \textbf{94.2\textsubscript{\tiny $\uparrow$9.4\%}} & \textbf{92.1\textsubscript{\tiny $\uparrow$6.2\%}} & \textbf{91.7\textsubscript{\tiny $\uparrow$6.5\%}} & \textbf{93.4\textsubscript{\tiny $\uparrow$8.2\%}} & \textbf{89.2\textsubscript{\tiny $\uparrow$8.0\%}} \\
\bottomrule
\end{tabular}%
}
\end{table*}

\section{Experiment}

To comprehensively assess the capabilities of our \textbf{HiVA} framework, we have conducted experiments across diverse task-driven environments. These experiments evaluate HiVA’s performance, efficiency, and the contributions of its key components through ablation studies, comparing it against state-of-the-art baselines.

\subsection{Experimental Setup}

We evaluated \textbf{HiVA} across mathematical reasoning, long-context multi-hop question answering, programmatic tasks, textual reasoning, and complex agentic environments, comparing it against state-of-the-art baselines. Ablation studies assessed the impact of HiVA’s components: Topological Evolution (TEV), Semantic Evolution (SEV), Knowledge-Aware Bayesian-Bandit Routing (KABB), environment feedback (Env), and tool integration (Tool). Adaptability and scalability experiments are conducted on the MBPP dataset to assess performance at different optimization steps. Experiments are conducted on both open-sourced models (\textit{e.g.}, \texttt{Qwen2.5-72B-Instruct-Turbo}) and close-sourced APIs (\textit{e.g.}, \texttt{GPT-4o-mini}).
All experiments are conducted with a fixed temperature of 1.0.

\subsubsection{Datasets}
We used the following benchmark datasets to cover diverse task domains: (1) Mathematical Reasoning: MATH~\cite{hendrycksmath2021} for complex problems and GSM-8K~\cite{cobbe2021gsm8k} for elementary problems; (2) Long-context Question Answering: HotpotQA~\cite{yang2018hotpotqa} and 2WikihopQA \cite{xanh2020_2wikimultihop} for multi-hop reasoning; (3) Programmatic Tasks: HumanEval~\cite{chen2021codex} and MBPP \cite{austin2021program} for code generation and problem-solving, with MBPP also used for adaptability and scalability experiments; (4) Textual Reasoning: MMLU~\cite{hendryckstest2021} for professional knowledge and BBH \cite{suzgun2022challenging} for challenging reasoning; and (5) Complex Agentic Environments: GAIA~\cite{mialon2024gaia} for interactive and agent-driven scenarios.

\subsubsection{Evaluation Metrics}
The performance is measured using accuracy (\%) averaged over five runs on randomly sampled data subsets. For the GAIA dataset, we evaluated performance using Accuracy (Acc) for task completion and Cost-efficiency Score (CS), calculated as Accuracy divided by the LLM's cost in dollars. Higher CS values indicate better cost-efficiency. The CS calculation method and sample strategy are detailed in the Appendix on Experimental Setup Details. 

\subsubsection{Baselines}
We compared HiVA against:
\textbf{CoT}~\cite{wei2022chain}: Step-by-step reasoning.
\textbf{Self-Consistency}~\cite{wang2023selfconsistency}: Majority voting over multiple reasoning paths.
\textbf{Self-Refine}~\cite{madaan2023self}: Iterative output refinement.
 \textbf{Multi-Agent Debate}~\cite{du2023improving}: Collaborative multi-agent reasoning.
 \textbf{DyLAN}~\cite{DBLP:journals/corr/abs-2310-02170}: Dynamic learning agent network.
\textbf{AgentVerse}~\cite{chen2023agentverse}: Multi-agent coordination framework.
 \textbf{ADAS}~\cite{hu2025automated}: Adaptive agent optimization.
\textbf{MaAS}~\cite{zhang2025agentic-supernet}: Multi-agent system with specialized roles.

\subsection{Main Results}

We evaluated HiVA and baseline methods across mathematical, long-context, programmatic, and textual reasoning tasks, as shown in Table~\ref{tab:task_performance}. The results highlight HiVA’s superior performance, achieving the highest average accuracy of 89.2\% (+8.0\% over Vanilla), with leading scores in GSM-8K (94.5\%, +3.6\%), HotpotQA (79.7\%, +18.3\%), 2WikihopQA (86.5\%, +13.5\%), MMLU (91.7\%, +6.5\%), and BBH (93.4\%, +8.2\%). HiVA’s semantic-topological evolution and knowledge-aware routing excel in tasks requiring multi-step reasoning and complex interactions, particularly in web-based and textual domains. Compared to MaAS (86.3\% average), which performs strongly in programmatic tasks (94.2\% on HumanEval, 90.1\% on MBPP), and ADAS (85.7\% average), which leads in MATH (86.1\%), HiVA demonstrates broader robustness. However, HiVA’s performance on MATH (81.2\%, -1.8\%) is slightly below Vanilla. As highlighted in our qualitative case study (Figure~\ref{fig:casestudy}), this performance drop can be attributed to the aggregator's difficulty in resolving conflicting answers generated by different agents during parallel verification. When faced with contradictory results, the aggregator can get ``stuck" and fail to produce a final answer, revealing a limitation in handling tasks that require strict logical consistency across multiple reasoning paths. Despite this, its strong results in MBPP (92.1\%, +6.2\%) and knowledge-intensive reasoning tasks underscore its effectiveness in handling other types of intricate dependencies.

\begin{figure}[ht]
    \centering
    \includegraphics[width=\linewidth]{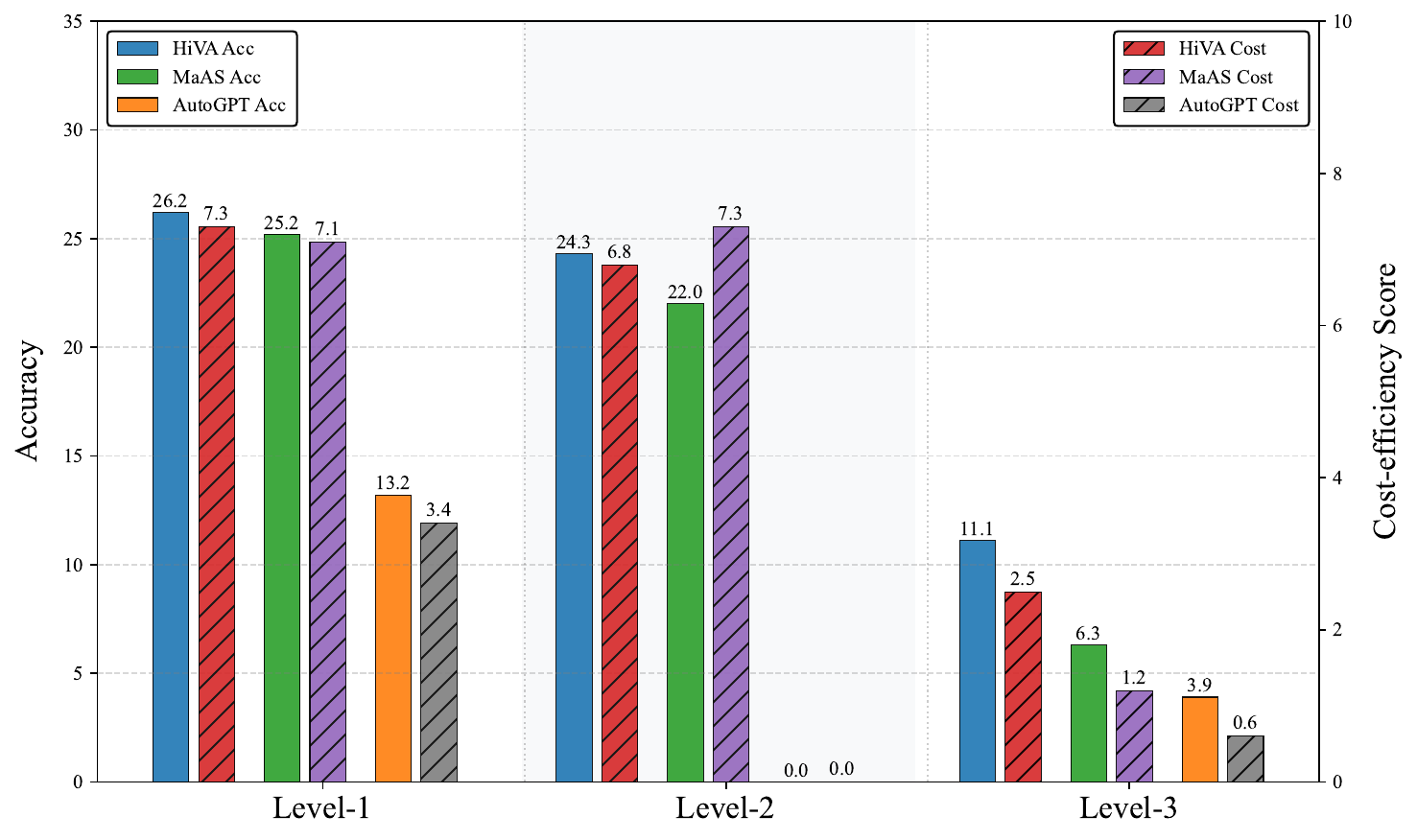}
    \caption{Evaluations in complex agentic environments. We compare two mainstream agentic frameworks (\textit{i.e.}, MaAS, AutoGPT) with HiVA in an open, complex benchmark (\textit{i.e.}, GAIA) and evaluate their performance through Accuracy (Acc) and Cost-efficiency Score (CS).}
    \label{fig:performance}
\end{figure}

\begin{figure*}[ht]
    \centering
    \includegraphics[width=0.9\linewidth]{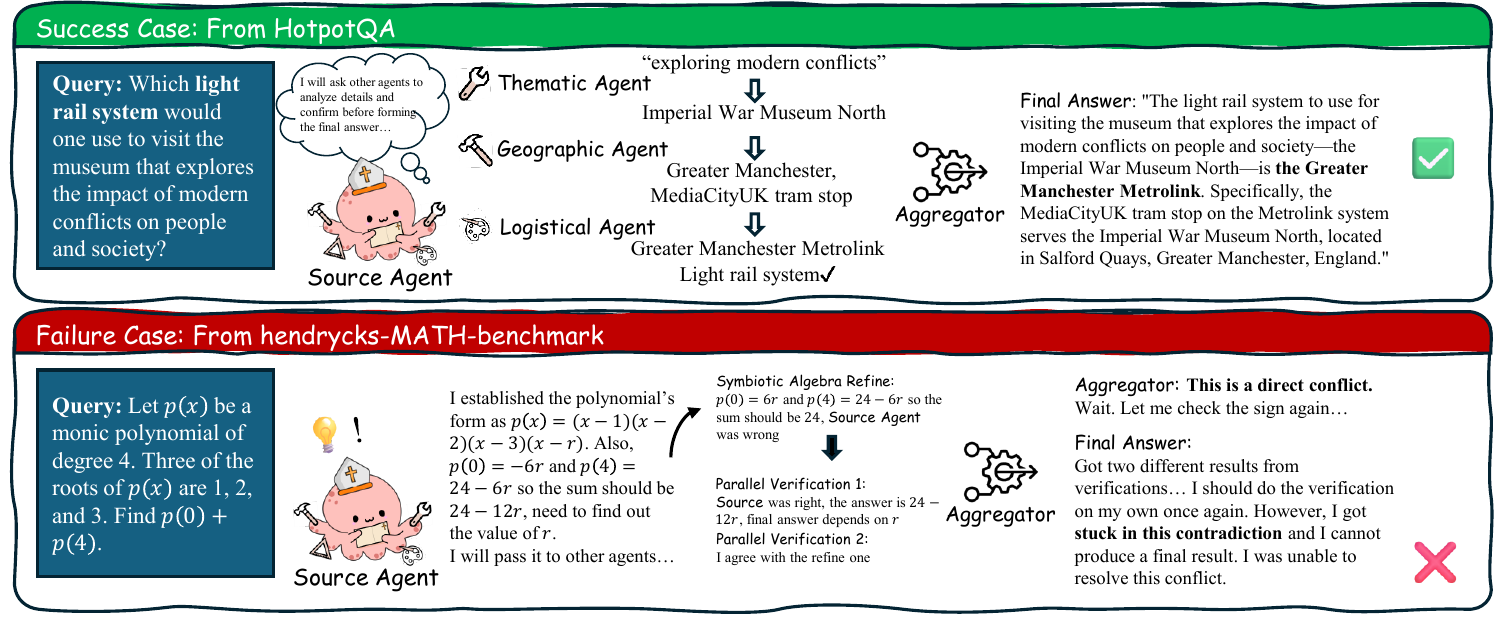}
    \caption{\textbf{Comparative Evolution Trajectories: Success vs. Failure Cases}. We evaluate our algorithm on HotpotQA and MATH, and choose two evolution trajectories (\textit{i.e.}, success and failure cases) from them.}
    \label{fig:casestudy}
\end{figure*}

In the complex agentic environment (GAIA), we assessed both Accuracy (Acc) and Cost-efficiency Score (CS), as shown in Figure~\ref{fig:performance}. HiVA consistently outperforms MaAS and AutoGPT in Acc across all task levels (Level-1: 26.2\%, Level-2: 24.3\%, Level-3: 11.1\% vs. MaAS: 25.2\%, 22.0\%, 6.3\% and AutoGPT: 13.21\%, 0.0\%, 3.9\%). The largest performance gap is in Level-2, where HiVA achieves 24.3\% Acc compared to MaAS’s 22.0\% and AutoGPT’s 0.0\%. Additionally, HiVA achieves the highest average CS (5.5) compared to MaAS (5.2) and AutoGPT (1.3), indicating superior cost-efficiency with fewer LLM calls. HiVA’s dynamic routing and adaptive agent coordination enable it to balance high accuracy with efficient resource use, outperforming workflow-based and reactive-loop-based methods in GAIA’s multi-step tasks. More experiments on optimization cost can be found in the Scalability Analysis section of the Appendix.

\begin{figure}[h]
    \centering
    \includegraphics[width=\linewidth]{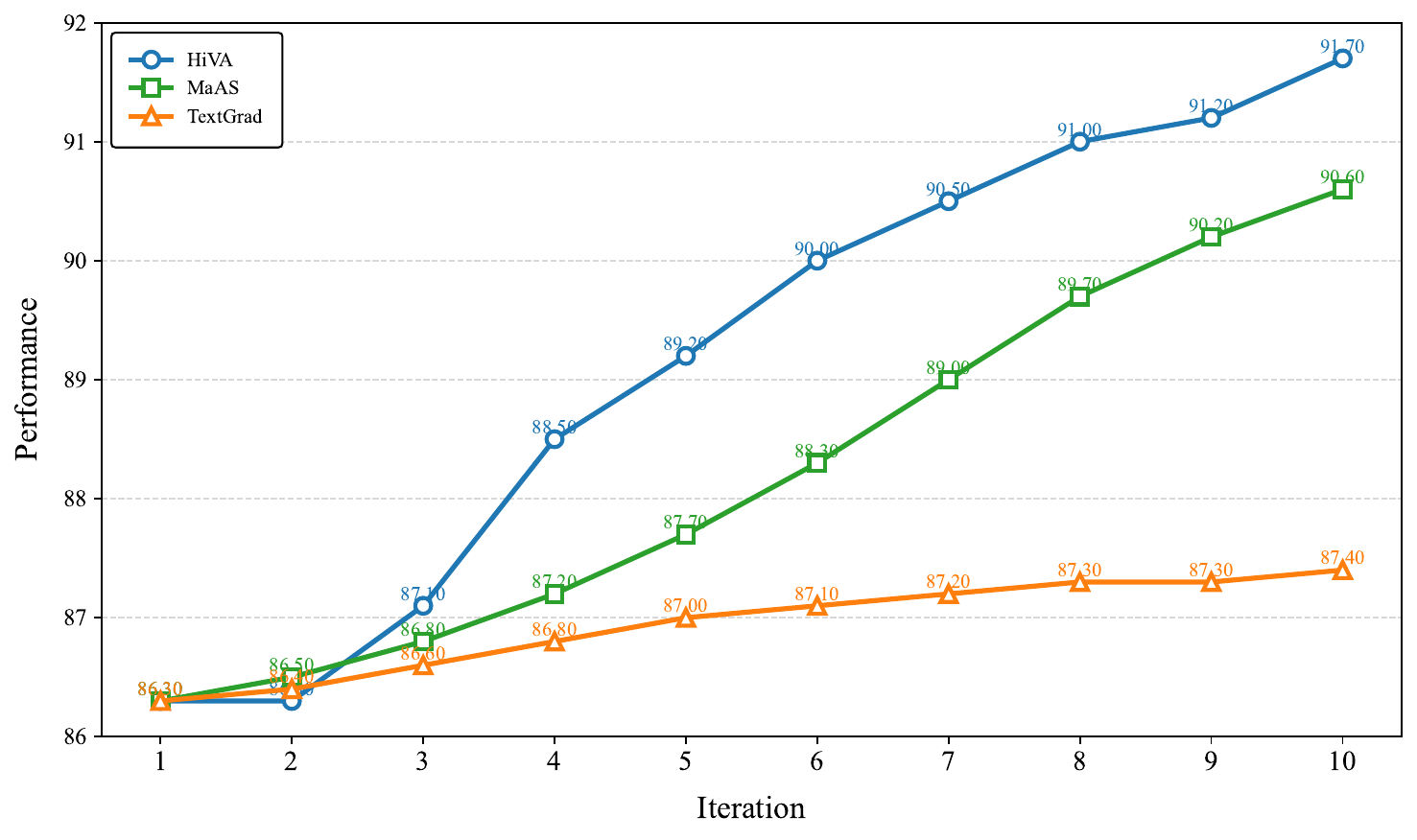}
    \caption{Adaptability and scalability trends of HiVA. Our HiVA is compared with MaAS and TextGrad across coding tasks in the MBPP of increasing iteration steps.}
    \label{fig:performance_vs}
\end{figure}

To evaluate HiVA’s adaptability and scalability, we conducted experiments on the MBPP dataset with tasks of increasing complexity (Introductory, Interview, Competition). Figure~\ref{fig:performance_vs} illustrates performance trends over 10 iterations for HiVA, MaAS, and TextGrad. HiVA demonstrates superior adaptability, improving from 86.3\% to 91.7\% (+5.4\%) over 10 iterations, surpassing MaAS (86.3\% to 90.6\%, +4.3\%) after iteration 4 and TextGrad (86.3\% to 87.4\%, +1.1\%), which plateaus early. HiVA’s steady performance gains, driven by its topological optimization and knowledge-aware routing, highlight its ability to scale effectively with task complexity while maintaining high accuracy.



\subsection{Qualitative Case Study}

To gain deeper insights into HiVA's behavior, we conducted a qualitative case study on representative tasks from each domain. For mathematical reasoning (MATH), we observed that HiVA's textual gradient mechanism effectively refines intermediate solutions, improving accuracy. In multi-hop question answering (HotpotQA), its knowledge-aware bandit-based routing dynamically selects relevant agents, enhancing multi-hop reasoning. For programmatic tasks (HumanEval), HiVA's topological optimization ensures efficient agent collaboration for robust code generation. In textual reasoning (MMLU), integrating environment feedback and tools enables precise knowledge retrieval. These observations highlight HiVA's ability to synergistically leverage its components across diverse tasks. Two illustrative cases, shown in Figure~\ref{fig:casestudy}, demonstrate HiVA's capabilities and limitations. More cases can be found in the Appendix on Qualitative Case Study.

The contrasting cases reveal critical factors determining HiVA's effectiveness:
 \textbf{Task Decomposability}: Multi-hop QA benefits from natural sequential decomposition, while complex bug fixing requires simultaneous reasoning across multiple interdependent components.
\textbf{Context Scalability}: Success correlates with tasks fitting within LLM context windows and agent specialization boundaries.
\textbf{Feedback Quality}: Clear, actionable environmental feedback enables effective textual gradient propagation, while ambiguous signals lead to local optimization traps.
\subsection{Ablation Studies}

\begin{table}[h]
\centering
\caption{Ablation Study of HiVA. Results show accuracy scores (\%) for each method across HotpotQA, programmatic (MBPP), and textual (MMLU) reasoning tasks. Bold indicates best performance. Subscript arrows denote relative change from full HiVA.}
\label{tab:ablation_study}
\resizebox{\linewidth}{!}{%
\begin{tabular}{@{}lcccr@{}}
\toprule
\textbf{Method} & \textbf{HotpotQA} & \textbf{MBPP} & \textbf{MMLU} & \textbf{Avg.} \\
\midrule
\textbf{HiVA (ours)} & \textbf{79.7} & \textbf{92.1} & \textbf{91.7} & \textbf{87.8} \\
HiVA w/o TEV & 74.0\textsubscript{↓7.3\%} & 88.9\textsubscript{↓3.5\%} & 88.3\textsubscript{↓3.7\%} & 83.7 \\
HiVA w/o SEV & 71.2\textsubscript{↓10.7\%} & 88.4\textsubscript{↓4.0\%} & 86.9\textsubscript{↓5.2\%} & 82.2 \\
HiVA w/o KABB & 76.2\textsubscript{↓4.4\%} & 88.1\textsubscript{↓4.4\%} & 90.6\textsubscript{↓1.2\%} & 85.0 \\
\midrule
HiVA w/o Env & 75.2\textsubscript{↓5.7\%} & 89.3\textsubscript{↓3.1\%} & 89.5\textsubscript{↓2.4\%} & 84.7 \\
HiVA w/o Tool & 74.8\textsubscript{↓6.1\%} & \underline{94.1\textsubscript{↑2.2\%}} & 89.1\textsubscript{↓2.8\%} & 84.3 \\
\bottomrule
\end{tabular}%
}
\end{table}

To understand the contribution of HiVA's key components, we conducted ablation studies by removing Topological Evolution (TEV), Semantic Evolution (SEV), Knowledge-Aware Bandit-Based Routing (KABB), environment feedback (Env), and tool integration (Tool). Table~\ref{tab:ablation_study} presents the results on HotpotQA, MBPP, and MMLU tasks.

The results validate our core hypothesis that both semantic and topological evolution are essential for adaptive intelligence. Semantic Evolution (SEV) proves most critical, with removal causing the largest degradation (10.7\% on HotpotQA, 5.2\% on MMLU), confirming that agents must learn ``what each agent should do." Topological Evolution (TEV) is equally important, particularly for multi-hop reasoning (7.3\% drop on HotpotQA), validating our claim that systems must evolve ``how agents should interact and organize."
The synergistic effect of TEV and SEV demonstrates our Semantic-Topological Evolution (STEV) algorithm. Neither component alone achieves optimal performance. Knowledge-Aware Bandit-Based Routing (KABB) consistently contributes (4.4\% decreases), confirming effective exploration of the evolved topological space. Environment feedback (Env) provides steady improvements, validating our textual gradient approach. More details can be found in the Appendix on Ablation Study Details.
\section{Conclusion}
We introduce HiVA, a framework for self-organized multi-agent systems that jointly evolves agent behaviors (semantics) and their collaboration structure (topology). Guided by environmental feedback, HiVA consistently outperforms baselines in accuracy and cost-efficiency on complex tasks like mathematical reasoning and code generation. Our work confirms that this co-evolution of semantics and topology is critical for optimal adaptation. Future work may develop more operable and effective tool-calling methods to better handle the challenges of dynamic environments.
\bibliography{aaai2026}

\newpage

\clearpage
\appendix

\section{Prompting Strategies}
\label{app:PromptingStrategies}

The operational core of HiVA's Semantic-Topological Evolution (STEV) algorithm is a set of structured prompt templates. These prompts serve as the concrete implementation that translates abstract optimization concepts—such as forward propagation along a dynamic graph, textual gradient-based feedback, and coordinated parameter updates—into executable instructions for the Large Language Models (LLMs) that constitute each agent. This section provides a detailed exposition of these core prompts, elucidating their design motivation, structure, and function within the HiVA framework.

\subsection{Core Prompt Templates for Agent Interaction and Evolution}
The interaction, feedback, and evolution loops in HiVA are orchestrated by a series of carefully designed prompts. Each prompt is engineered to elicit a specific cognitive task from the LLM, ranging from instruction formulation to structural network optimization.

\subsubsection{Forward Propagation}
In the forward pass, HiVA dynamically constructs an execution subgraph tailored to the specific task. For this graph to be computationally effective, information must flow coherently between connected nodes (agents). The motivation behind the successor instruction prompt is to create a robust mechanism for this information transfer. The prompt detailed in Figure~\ref{fig:forward_prompt} is the key mechanism for generating the ``message" that travels along an edge of the graph, effectively translating the output of a predecessor agent into a relevant, actionable input for its successor.

\begin{figure*}[!t]
\begin{tcolorbox}[colback=blue!5!white,colframe=blue!75!black,title=Forward Propagation: Successor Instruction,width=\textwidth]
\textbf{Objective:} Generate instructions for successor agent $j$ from agent $i$'s input and output, and $j$'s system prompt. \\
\textbf{Template:} \\
You are an instruction generator for multi-agent systems. Create clear, specific instructions. \\
\textbf{User Prompt:} \\
Generate an instruction for the successor agent: \\
CURRENT AGENT INPUT: \texttt{\{input\_instruction\}} \\
CURRENT AGENT TOOL RESULT: \texttt{\{tool\_result\}} \\
SUCCESSOR AGENT SYSTEM PROMPT: \texttt{\{successor.system\_prompt\}} \\
SUCCESSOR AGENT ID: \texttt{\{successor.agent\_id\}} \\
Requirements: 1. Actionable instruction. 2. Align with the successor’s capabilities. 3. Transfer relevant context. 4. Be concise. \\
\textbf{Example:} \\
Input: ``Population of Tesla’s headquarters city?" \\
Tool Result: ``Tesla Inc. headquarters: Austin, Texas." \\
Successor Prompt: ``You are a data retrieval agent." \\
Output: ``Retrieve Austin, Texas population."
\end{tcolorbox}
\caption{Forward propagation prompt for generating successor instructions, ensuring context transfer between agents.}
\label{fig:forward_prompt}
\end{figure*}

\subsubsection{Backward Propagation}
The backward pass is central to HiVA's learning capability. A key challenge is converting feedback from the external environment, which is often a black-box signal, into a usable optimization signal. Since the agentic graph is discrete and non-differentiable, HiVA employs the concept of a ``textual gradient." This process is operationalized in two stages.

First, a global textual gradient is generated from the raw environmental feedback. The prompt shown in Figure~\ref{fig:backward_system_prompt} tasks the aggregator agent with this diagnostic role, creating a high-level error signal that identifies the primary deficiency in the final output. Subsequently, this global gradient must be localized for agent-specific updates. The prompt in Figure~\ref{fig:backward_agent_prompt} performs this crucial credit assignment. It decomposes the feedback received by successors and attributes responsibility to the current agent's output, generating a localized gradient with specific critiques for the agent's prompt and tool.

\begin{figure*}[!t]
\begin{tcolorbox}[colback=blue!5!white,colframe=blue!75!black,title=Backward Propagation: System Feedback,width=\textwidth]
\textbf{Objective:} Generate system feedback from environmental feedback and agent outputs. \\
\textbf{Template:} \\
You are an output aggregator generating feedback for predecessors based on environmental outcomes. \\
\textbf{User Prompt:} \\
Generate feedback for: \\
PREDECESSOR: \texttt{predecessor agents} \\
SUCCESSOR: \texttt{Aggregator \{aggregator\_id\}} \\
SUCCESSOR FEEDBACK: \texttt{\{loss\_grad\}} \\
CONTEXT: \texttt{Final result: \{final\_result[:200]\}...} \\
Output: \texttt{<FEEDBACK>\{feedback\}</FEEDBACK>} \\
\textbf{Example:} \\
Loss Gradient: ``Inaccurate population." \\
Final Result: ``Austin population: 950,000." \\
Output: \texttt{<FEEDBACK>Use recent sources for Austin’s population.</FEEDBACK>}
\end{tcolorbox}
\caption{Backward propagation prompt for generating system-level feedback based on environmental outcomes.}
\label{fig:backward_system_prompt}
\end{figure*}

\begin{figure*}[!t]
\begin{tcolorbox}[colback=blue!5!white,colframe=blue!75!black,title=Backward Propagation: Agent Feedback,width=\textwidth]
\textbf{Objective:} Generate feedback for agent $i$ from successors’ feedback and $i$'s output. \\
\textbf{Template:} \\
Analyze successor feedback for agent improvement. \\
\textbf{User Prompt:} \\
Current Agent Role: \texttt{\{system\_prompt[:300]\}...} \\
Successor Feedback: \texttt{\{combined\_feedback\}} \\
Provide: \\
1. SYSTEM\_PROMPT\_FEEDBACK: Role improvements. \\
2. TOOL\_FEEDBACK: Tool improvements. \\
3. OVERALL\_FEEDBACK: Strategic improvements. \\
\textbf{Example:} \\
Feedback: ``Agent A1: Missing entities." \\
Output: \\
SYSTEM\_PROMPT\_FEEDBACK: Define entity types. \\
TOOL\_FEEDBACK: Enhance entity extraction tool. \\
OVERALL\_FEEDBACK: Improve context transfer.
\end{tcolorbox}
\caption{Backward propagation prompt for generating agent-specific feedback from successor agents.}
\label{fig:backward_agent_prompt}
\end{figure*}

\subsubsection{Semantic and Topological Evolution}
Once localized textual gradients are generated, the final step is to apply them to evolve the multi-agent system. HiVA's core innovation is the co-evolution of both agent semantics and network topology. The prompt in Figure~\ref{fig:prompt_update} operationalizes semantic evolution ($f_P$) by instructing the LLM to rewrite an agent's parameters (\textit{e.g.}, its prompt) to incorporate the received feedback. This process is analogous to a text-based gradient descent step. Complementing this, the prompt in Figure~\ref{fig:topology_prompt} drives topological evolution ($f_G$). It frames the network modification as a structured decision-making task, allowing an agent to adapt its local connectivity based on performance and task structure.

\begin{figure*}[!t]
\begin{tcolorbox}[colback=blue!5!white,colframe=blue!75!black,title=Prompt Update Generation,width=\textwidth]
\textbf{Objective:} Update system prompt and tool function based on feedback. \\
\textbf{Template:} \\
Generate an improved system prompt. \\
\textbf{User Prompt:} \\
Current variable: \texttt{\{system\_prompt\}} \\
Role: \texttt{AI agent system prompt} \\
Gradients: \texttt{\{feedback['system\_prompt\_feedback']\}} \\
Output: \texttt{<IMPROVED\_VARIABLE>\{prompt\}</IMPROVED\_VARIABLE>} \\
\textbf{Example:} \\
Current Prompt: ``You are a reasoning agent." \\
Feedback: ``Specify entity extraction." \\
Output: \texttt{<IMPROVED\_VARIABLE>You are a reasoning agent specializing in extracting entities (\textit{e.g.}, locations).</IMPROVED\_VARIABLE>}
\end{tcolorbox}
\caption{Prompt for generating updated system prompts based on textual gradients from feedback.}
\label{fig:prompt_update}
\end{figure*}

\begin{figure*}[!t]
\begin{tcolorbox}[colback=blue!5!white,colframe=blue!75!black,title=Topological Decision Prompt,width=\textwidth]
\textbf{Objective:} Decide agent $i$'s local topology changes. \\
\textbf{Template:} \\
You are a network topology optimizer. \\
\textbf{User Prompt:} \\
Current Agent Role: \texttt{\{system\_prompt\}...} \\
Feedback: \texttt{\{feedback\}} \\
Successor Count: \texttt{\{len(successors)\}} \\
Task Parallelizability: \texttt{\{parallelizability\}} \\
Options: \\
- ADD\_PARALLEL: [New agent description] \\
- ADD\_SERIAL: [New agent description] \\
- REMOVE\_SUCCESSOR: [Successor to remove] \\
- NO\_CHANGE: [Reason] \\
\textbf{Example:} \\
Feedback: ``Needs parallel subtasks." \\
Output: \texttt{ADD\_PARALLEL: Data retrieval agent}
\end{tcolorbox}
\caption{Prompt for deciding local topology changes, such as adding or removing successor agents.}
\label{fig:topology_prompt}
\end{figure*}

\subsection{Prompt Evolution Examples}
To render the abstract process more concrete, Figure~\ref{fig:prompt_evolution} illustrates a complete, single-iteration trajectory of semantic evolution. It demonstrates how a generic `Reasoning Agent' receives a targeted textual gradient and subsequently refines its system prompt to be more specific and robust, leading to a higher-quality output. This case highlights the direct impact of the feedback loop on agent behavior.

\begin{figure*}[!t]
\begin{tcolorbox}[colback=green!5!white,colframe=green!75!black,title=Prompt Evolution: Reasoning Agent,width=\textwidth]
\textbf{Task:} ``Which light rail serves the Imperial War Museum North?" \\
\textbf{Initial Prompt:} \\
You are a reasoning agent. Decompose task: \texttt{\{task\_description\}} into steps. Output in markdown. \\
\textbf{Initial Output:} \\
```markdown
1. Identify museum location.
2. Find the light rail system.
``` \\
\textbf{Textual Gradient:} \\
SYSTEM\_PROMPT\_FEEDBACK: Cross-reference documents for accuracy. \\
\textbf{Evolved Prompt:} \\
\texttt{<IMPROVED\_VARIABLE>You are a reasoning agent. Decompose task: \{task\_description\} into steps, cross-referencing documents for accuracy. Output in markdown.</IMPROVED\_VARIABLE>} \\
\textbf{Evolved Output:} \\
```markdown
1. Cross-reference: Imperial War Museum North in Manchester, UK.
2. Identify light rail: Greater Manchester Metrolink, MediaCityUK stop.
``` \\
\textbf{Result:} \textcolor{green!70!black}{SUCCESS}
\end{tcolorbox}
\caption{Evolution of a Reasoning Agent’s prompt, showing improvement driven by textual gradients.}
\label{fig:prompt_evolution}
\end{figure*}

\subsection{Evolvable Tool Subsystem}
In the HiVA framework, tools are not static plugins but are treated as first-class, evolvable components of an agent's semantic identity. An agent's effectiveness is determined by both its intent (system prompt) and its capabilities (equipped tools). Consequently, we have designed a comprehensive tool subsystem that supports the entire lifecycle of a tool: from definition and secure execution to dynamic generation and feedback-driven evolution. This subsystem is architected upon the principles of structured representation, governed execution, and adaptive evolution.

\subsubsection{Structured Representation and Governed Execution}
The foundation of our tool subsystem is the principle that every tool is formalized as a structured schema rather than an opaque function. This schema encapsulates essential metadata, including a natural-language description of the tool's purpose, a formal definition of its input/output parameters, and explicit safety or operational constraints. A central registry maintains these schemas, tracking versioning and performance metrics.

When a tool is invoked, its execution is managed by a governed runtime environment that enforces operational policies (\textit{e.g.}, timeouts, retries) and performs security validation to prevent unsafe operations. This robust process ensures that even dynamically evolved tools adhere to strict safety protocols, guaranteeing the stability of the multi-agent system.

\subsubsection{Adaptive Evolution through Synthesis and Refinement}
The subsystem's most advanced capability is the adaptive evolution of tools, which is realized through a dual mechanism of synthesis and refinement. This allows the multi-agent system to autonomously expand and improve its functional repertoire.

\begin{itemize}
    \item \textbf{De Novo Synthesis (Tool Generation):} The framework can synthesize entirely new tools from a high-level capability description. This generative process, driven by the prompt shown in Figure~\ref{fig:tool_synthesis_prompt}, allows the system to create novel functionalities on demand to address unforeseen problems.
    
    \item \textbf{Iterative Refinement (Tool Update):} Existing tools undergo continuous improvement. Performance failures or inefficiencies generate a `TOOL\_FEEDBACK' gradient. This textual gradient, along with the tool's current source code, is used to operationalize the refinement process. The prompt detailed in Figure~\ref{fig:tool_refinement_prompt} guides a code-synthesis LLM to rewrite the tool's logic, patching bugs or adding functionality in response to the feedback.
\end{itemize}

Through these mechanisms, the toolset of the agent collective is not a fixed asset but a dynamic, constantly improving component of the system's intelligence.

\begin{figure*}[!t]
\begin{tcolorbox}[colback=blue!5!white,colframe=blue!75!black,title=Tool Synthesis Prompt: Generating New Capabilities]
\textbf{Objective:} To generate a secure and valid Python function from a natural language description, enabling the system to create new tools on-the-fly.

\textbf{Template:} \\
You are a tool generation specialist for an AI agent system. Your task is to write a single, self-contained Python function based on the provided requirements.

\textbf{User Prompt:} \\
Generate a Python tool function with the following specifications:

**Tool Description:**
\texttt{\{A natural language description of the tool's purpose, \textit{e.g.}, ``A function to calculate the SHA256 hash of a given string."\}}

**Input/Output Examples:**
\texttt{\{Optional examples, \textit{e.g.}, ``Input: 'hello world', Output: 'b94d27b9934d3e08a52e52d7da7dabfac484efe37a5380ee9088f7ace2efcde9'"\}}

**Function Signature Requirements:**\\
- The function must be named \texttt{tool\_function}.\\
- It must accept at least one parameter: \texttt{input\_data}.\\
- It must include error handling for invalid inputs.\\
- It must not use restricted libraries like \texttt{os} or \texttt{subprocess}.

Please return only the complete Python code inside a \texttt{python} block.

\textbf{Example of Expected Output (for SHA256 hash):}
\begin{lstlisting}[language=Python]
import hashlib

def tool_function(input_data: str, **kwargs):
    """
    Calculates the SHA256 hash of a given string.
    """
    if not isinstance(input_data, str):
        return "Error: Input must be a string."
    try:
        # Encode the string to bytes before hashing
        encoded_string = input_data.encode('utf-8')
        sha256_hash = hashlib.sha256(encoded_string).hexdigest()
        return sha256_hash
    except Exception as e:
        return f"Error during hashing: {str(e)}"
\end{lstlisting}
\end{tcolorbox}
\caption{The prompt template used by the `ToolGenerator' to synthesize a new tool from a high-level description. This enables the agent system to dynamically acquire new skills.}
\label{fig:tool_synthesis_prompt}
\end{figure*}

\begin{figure*}[!t]
\begin{tcolorbox}[colback=green!5!white,colframe=green!75!black,title=Tool Refinement Prompt: Evolving Existing Capabilities]
\textbf{Objective:} To rewrite an existing tool's source code to address specific critiques from performance feedback, thereby iteratively improving its robustness and functionality.

\textbf{Template:} \\
You are a tool evolution specialist. Your task is to improve an existing Python tool function based on its source code and performance feedback.

\textbf{User Prompt:} \\
Please improve the following tool function.

**Current Tool Function Source Code:**\texttt{\{Source code of the current tool\}}

**Performance Feedback (Textual Gradient):**
\texttt{\{The textual gradient describing the problem.\}}

**Improvement Requirements:**\\
- Preserve the original function signature (\texttt{tool\_function(input\_data, **kwargs)}).\\
- Directly address the issue described in the feedback.\\
- Enhance error handling and edge-case coverage.\\
- Do not introduce new external dependencies.
Return only the improved, complete Python function code inside a \texttt{python} block.

\textbf{Example (for a simple calculator tool):}\\
    - \textbf{Current Source Code}:
    \begin{lstlisting}[language=Python, firstnumber=1]
import re
def tool_function(input_data: str, **kwargs):
    # Extracts two numbers and calculates their ratio.
    numbers = re.findall(r'\d+\.?\d*', input_data)
    a, b = float(numbers[0]), float(numbers[1])
    return f"The ratio is: {a / b}"
    \end{lstlisting}
    
    - \textbf{Feedback}: ` ``Tool execution failed with a ZeroDivisionError. The logic must be updated to handle cases where the denominator is zero."' 
    
    - \textbf{Expected Evolved Code}:
    \begin{lstlisting}[language=Python, firstnumber=1]
import re
def tool_function(input_data: str, **kwargs):
    # Extracts two numbers and calculates their ratio safely.
    numbers = re.findall(r'\d+\.?\d*', input_data)
    a, b = float(numbers[0]), float(numbers[1])
    if b == 0:
        return "Error: Cannot divide by zero."
    return f"The ratio is: {a / b}"
    \end{lstlisting}
\end{tcolorbox}
\caption{The prompt template used by the `ToolUpdater' to evolve an existing tool. This process treats the tool's code as a mutable parameter that is optimized using textual gradients, directly realizing the concept of tool evolution.}
\label{fig:tool_refinement_prompt}
\end{figure*}

\section{Experimental Setup Details}

This section provides comprehensive details on the experimental setup for evaluating the HiVA framework, ensuring reproducibility and transparency.

\subsection{Random Sampling Methodology}
To ensure a fair and comprehensive evaluation across a wide range of tasks while maintaining computational tractability, we employed a rigorous random sampling methodology. For each benchmark, we sampled from the official test or validation sets. All sampling procedures were performed using NumPy's programmatic functions with a fixed random seed (\texttt{seed=42}) to guarantee the reproducibility of our test partitions. The specific strategy for each benchmark is detailed below:

\begin{itemize}
    \item \textbf{Mathematical Reasoning (MATH \& GSM-8K):} For the \textbf{MATH} dataset, which covers a wide range of difficulties, we performed stratified sampling to select 500 problems from the 12,000-problem test set. The stratification ensures that the proportion of problems from each subject (\textit{e.g.}, Prealgebra, Algebra, Geometry, etc.) in our sample mirrors that of the original dataset. For the simpler \textbf{GSM-8K} dataset, we randomly sampled 500 problems from its official test set.

    \item \textbf{Long-Context QA (HotpotQA \& 2WikihopQA):} For \textbf{HotpotQA}, known for its multi-hop reasoning requirements, we sampled 200 questions from the development set, ensuring an equal representation of ``bridge" and ``comparison" question types through stratification. Similarly, for \textbf{2WikihopQA}, we randomly sampled 200 questions from its validation set.

    \item \textbf{Programmatic Tasks (HumanEval \& MBPP):} For \textbf{HumanEval}, we randomly sampled 50 distinct coding tasks from the full set of 164 problems. For the \textbf{MBPP} dataset, which is also used in our scalability analysis, we sampled 100 problems from its test set, stratified across the documented difficulty levels to ensure a balanced evaluation.

    \item \textbf{Textual Reasoning (MMLU \& BBH):} Evaluating on the massive \textbf{MMLU} benchmark required careful sampling. We constructed a 500-question subset by performing stratified sampling across its 57 subjects. The number of questions sampled from each subject was proportional to its representation in the original test set, ensuring broad domain coverage. For the \textbf{BBH} (BIG-Bench Hard) suite, we sampled 20 instances from each of its 27 challenging sub-tasks, resulting in a test bed of 540 instances.

    \item \textbf{Complex Agentic Environments (GAIA):} Due to the small size and high complexity of the \textbf{GAIA} benchmark's test set, we did not perform random sampling. Instead, we conducted a full evaluation on all available Level 1, 2, and 3 tasks to provide a complete and definitive assessment of performance in these complex, multi-step scenarios.
\end{itemize}

\subsection{Cost-efficiency Score Calculation}
The cost-efficiency score balances performance and computational cost. For a task \( t \), let \( P(t) \in [0, 1] \) be the performance (\textit{e.g.}, accuracy for MATH, pass@1 for HumanEval), and \( C(t) \) be the cost in seconds of inference time. The score is defined as:
\[
S(t) = \frac{P(t) \times 100}{C(t) + \epsilon},
\]
where \(\epsilon = 0.01\) prevents division by zero. Cost \( C(t) \) is estimated as the total inference cost in dollars across all agents, measured on the Together.AI API.

\subsection{Hyperparameters}
HiVA's operation is governed by a set of hyperparameters spanning agent, system, and tool configurations. For agent-level parameters, instruction generation prompts use a temperature of 1.0 and a maximum of 1000 tokens, while feedback is weighted with 0.7 for system-level changes and 0.3 for tool modifications. System-wide settings include a parallelizability threshold of 0.5 for topology decisions, a maximum of 5 successors per agent, and a robust tool execution policy with a 30-second timeout and 3 retries. Tool-specific parameters for code generation are set to a temperature of 0.3 and a maximum of 1000 tokens, with all dynamic code executed in a code sandbox to ensure stability.

\subsection{Reproducibility Details}
Experiments are conducted on a cluster with 8 NVIDIA A100 GPUs, 1024GB RAM, and Ubuntu 22.04. Software included Python 3.10, PyTorch 2.10, and Transformers 4.38.0. The LLM backbone is \texttt{Qwen-2.5-72B-Instruct}. Code is available at \url{https://anonymous.4open.science/r/HiVA-60C6} with seed 42 for all random operations.

\section{Scalability Analysis}

While the HiVA framework demonstrates significant adaptive capabilities through its semantic-topological evolution, the computational overhead of the evolution process itself is an important consideration and represents a minor limitation of our approach.

The optimization cost is primarily driven by the multiple Large Language Model (LLM) calls required within each optimization iteration. These calls are utilized for the forward pass, which includes dynamic routing and instruction generation; the backward pass for generating textual gradients from environmental feedback; and the final coordinated updates to agent semantics and topology. Consequently, the optimization cost scales with the number of optimization iterations and the size of the multi-agent network (i.e., the number of agents, $|\mathcal{V}|$).

Theoretically, the worst-case time complexity for a single optimization iteration is $O(|\mathcal{V}|^2)$. This suggests a potential for exponential growth in cost as the agent network becomes larger and more complex. However, HiVA effectively mitigates this issue in practice through its core design principles:

\begin{itemize}
    \item \textbf{Knowledge-Aware Dynamic Routing (KABB):} Instead of activating all agents at each step, the KABB mechanism dynamically constructs a sparse, task-specific execution subgraph. This significantly reduces the number of LLM calls required for a single forward pass, avoiding the ``broadcast" overhead typical in larger networks.
    \item \textbf{Topological Pruning:} The \texttt{RepairTopology} function periodically removes inefficient or redundant connections based on historical performance. This ensures that the network topology remains relatively sparse and efficient as it evolves, thereby controlling the growth in complexity.
\end{itemize}

Thanks to these mechanisms, despite the theoretical complexity, the overall API overhead of HiVA remains competitive when compared to other multi-agent methods that might rely on extensive trial-and-error or exhaustive debate rounds.

To provide a concrete cost metric, we analyzed the default settings used in our experiments (\textit{e.g.}, 10 optimization iterations as demonstrated in Figure 5). When using the \texttt{Qwen-2.5-72B-Instruct} model, the average API consumption to perform the full optimization process for a single sample on the GAIA benchmark is approximately \textbf{\$0.1}. This indicates that while the optimization cost is a valid concern, the cost-benefit profile is reasonable given the significant performance improvements that HiVA delivers.

To provide a concrete cost metric, we analyzed the default settings used in our experiments (\textit{e.g.}, 10 optimization iterations as demonstrated in Figure 5). When using the \texttt{Qwen-2.5-72B-Instruct} model, the average API consumption to perform the full optimization process for a single sample on the GAIA benchmark is approximately \textbf{\$0.1}. This indicates that while the optimization cost is a valid concern, the cost-benefit profile is reasonable given the significant performance improvements that HiVA delivers.

\section{Knowledge-Based Cost Function}
\label{app:KnowledgeBasedCostFunction}

\subsection{Conceptual Framework}
A core challenge in dynamic multi-agent systems is efficiently routing tasks to the most suitable agents. Traditional Multi-Armed Bandit (MAB) approaches often rely solely on historical performance feedback, overlooking the crucial semantic relationships between tasks and agent capabilities. To address this gap, inspired by recent work in knowledge-aware coordination~\cite{zhang2025kabb}, our HiVA framework incorporates a Knowledge-Based Cost Function, $Dist(A_i, I_{\text{task}})$. Unlike approaches that evaluate the synergy of an entire expert team, our function is designed to assess the fitness of an individual agent $A_i$ for a given task $I_{\text{task}}$. This focus on individual assessment is crucial for the iterative, agent-by-agent construction and refinement of the execution graph in HiVA. The function serves as a sophisticated penalty term within the MAB selection process, providing a principled basis for routing novel tasks for which no performance history exists. It is formulated as:
\[
Dist(A_i, I_{\text{task}}) = \log(1 + d_I) \cdot \sum_{k=1}^{4} \omega_k \Psi_k
\]
This formulation is composed of a \textbf{Task Complexity Factor}, $\log(1 + d_I)$, which scales the penalty based on the task's intrinsic difficulty, and a \textbf{Weighted Mismatch Score}, a linear combination of four distinct mismatch indicators ($\Psi_k$) derived from a domain knowledge graph.

\subsection{Hyperparameter Settings}
The parameters governing the KABB-inspired routing mechanism are calibrated based on empirical analysis from the source work~\cite{zhang2025kabb} to ensure optimal performance. The specific values and descriptions are as follows:
\begin{itemize}
    \item \textbf{Knowledge Distance Threshold:} The system achieves optimal performance when this threshold is set to \textbf{0.75}. A lower threshold tends to include irrelevant experts, while a higher threshold makes the selection too restrictive, impacting system efficiency and coverage~\cite{zhang2025kabb}.
    \item \textbf{Time Decay Factor:} The optimal value for the time decay factor is set to \textbf{0.6}. This value strikes a balance between leveraging historical performance data and adapting to recent changes in agent effectiveness. A smaller factor makes the system overly sensitive to short-term fluctuations, whereas a larger one can suppress adaptability~\cite{zhang2025kabb}. This factor is used to calculate the term $\gamma^{\Delta t} = e^{-\kappa \Delta t}$.
    \item \textbf{Indicator Weights ($\omega_k$):} The weights $\omega_k$ for the four sub-indicators in the cost function are treated as learnable parameters that satisfy the constraint $\sum \omega_k = 1$. They are optimized on a held-out validation set to maximize the correlation between the cost function and actual task success, rather than being fixed constants~\cite{zhang2025kabb}.
    \item \textbf{Other Learnable Parameters:} Other key hyperparameters, such as the knowledge distance penalty factor ($\lambda$), the team synergy influence ($\eta$), and the knowledge matching correction strength ($\delta$), are also treated as part of the model's learnable configuration, optimized for the specific task domain rather than being pre-set to a single value.
\end{itemize}

\subsection{Knowledge Graph Architecture}
The computation of the cost function is grounded in a domain-specific Knowledge Graph (KG), which serves as the structured representation of the problem space and the system's capabilities. Its architecture is defined by three primary components: (1) \textbf{Nodes}, which are heterogeneous and represent Concepts (\textit{e.g.}, ``Algebra", ``Python Programming"), Agents (\textit{e.g.}, $A_i$), and Tools (\textit{e.g.}, ``Code Interpreter"); (2) \textbf{Edges}, which are typed and directed to capture semantic relationships like \texttt{is\_a}, \texttt{requires\_skill}, and \texttt{has\_tool}; and (3) \textbf{Construction and Embeddings}, a process where the graph is semi-automatically built from domain-specific corpora and each node is associated with a vector embedding to enable fine-grained semantic similarity calculations.

\subsection{Reward Signal Design}

In the KABB update formulas, the reward signal $r_i^{(t)}$ serves as a critical learning component that enables the routing policy to adapt based on performance history. We define this mechanism through a sophisticated credit assignment process that transforms sparse global feedback into granular agent-specific rewards.

The reward signal $r_i^{(t)}$ operates as a binary indicator where $r_{i}^{(t)} \in \{0, 1\}$, derived from global task outcomes through localized credit assignment. This process unfolds across three key stages.

\subsubsection{Global Outcome Collection}
The system initially receives sparse, environment-level feedback from $\mathcal{E}_{env}$ based on final task completion status. Successful task execution generates a global success signal, while failures produce corresponding failure indicators.

\subsubsection{Textual Gradient-Based Credit Assignment}
Since global signals provide insufficient granularity for individual agent updates, HiVA employs textual gradient backpropagation to distribute credit appropriately. During the backward pass, each participating agent $v_i$ receives a localized textual gradient ($\frac{\partial\mathcal{L}_{t}}{\partial v_{i}}$) containing detailed feedback about its specific contribution to the overall task performance.

\subsubsection{Sentiment-Based Reward Generation}
The localized textual gradients undergo sentiment analysis through an LLM to determine individual rewards $r_i^{(t)}$. Positive or neutral feedback (such as ``The financial data was extracted correctly") indicates successful sub-task completion, resulting in $r_i^{(t)} = 1$. Critical feedback (like ``The agent failed to handle the edge case, causing a runtime error") leads to $r_i^{(t)} = 0$.

This architecture effectively decomposes sparse global success signals into nuanced, agent-specific feedback. The resulting granularity enables the KABB routing policy to learn with enhanced precision, appropriately rewarding agents that contributed positively to task outcomes, even when overall execution fails due to errors from other agents in the system.

\subsection{Cost Component Definitions}

\subsubsection{Task Complexity Factor ($d_I$)}
The term $d_I$ quantifies the specificity of the task instruction $I_{\text{task}}$. Drawing inspiration from topological depth in knowledge graphs, we define $d_I$ as the shortest path distance in the KG from a general root ``Task" node to the primary concept node associated with the instruction. A more specialized task will have a greater depth and thus a larger $d_I$. The logarithmic scaling, $\log(1 + d_I)$, ensures that the mismatch penalty is amplified for more complex tasks where selecting the correct specialist is critical.

\subsubsection{Mismatch Sub-indicators ($\Psi_k$)}
The four sub-indicators were chosen to provide a holistic measure of agent-task fitness, capturing distinct dimensions of compatibility. Each indicator $\Psi_k$ is normalized to $[0, 1]$: (i) \textbf{Semantic Mismatch} ($\Psi_1$) measures the semantic distance between an agent's expertise and the task's requirements, extending prior work's Jaccard-based overlap metric by also incorporating the cosine distance between node embeddings to capture finer semantic nuances; (ii) \textbf{Dependency Complexity} ($\Psi_2$) measures the structural effort required to apply an agent's skills to the task and is calculated as the normalized shortest dependency path in the KG; (iii) \textbf{Historical Performance Mismatch} ($\Psi_3$) refines the use of historical data by considering an agent's success rate only on semantically similar past tasks, making the metric more context-aware; and (iv) \textbf{Tool Incompatibility} ($\Psi_4$), which provides a strong penalty for functional mismatches, a metric specifically adapted for our individual-agent assessment in place of team-based synergy.

\subsubsection{Indicator Weights ($\omega_k$)}
The weights $\omega_k$ balance the contribution of each indicator and satisfy $\sum \omega_k = 1$. They are treated as learnable parameters, optimized on a held-out validation set of tasks to maximize the correlation between the overall cost function and actual task success.

\section{Qualitative Case Studies}
To provide a granular view of HiVA's dynamic operational capabilities, this section presents qualitative analyses of its behavior on complex tasks. These case studies go beyond quantitative metrics to illustrate the framework's core mechanisms, including autonomous task decomposition, agent specialization, and the process of semantic-topological evolution in response to environmental feedback.

\subsection{Case Study 1: Successful Self-Organization and Evolution}
We present HiVA with a complex research task that requires synthesizing information from multiple modalities: \textit{``Analyze the Q2 2025 earnings report for 'TechCorp', cross-reference market sentiment from financial news articles in the same period, and generate a summary of key risks and opportunities."}

Initially, at iteration $t=0$, the HiVA network consists of a single, general-purpose agent. This agent attempts the task but produces a superficial summary, as it lacks the specialized tools and procedures to parse financial documents and gauge market sentiment. The environmental feedback is consequently negative, yielding a textual gradient criticizing the output for its ``lack of specific financial metrics and absence of market context."

This pointed feedback triggers the evolution process. In the backward pass, HiVA's topological evolution function ($f_G$) hypothesizes that the task is decomposable. It instigates a structural change, splitting the single generalist agent into two new, specialized agents: \textbf{$A_{\text{Financial\_Analyzer}}$}, with a system prompt geared towards extracting key metrics (\textit{e.g.}, revenue, EPS) from structured documents, and \textbf{$A_{\text{Sentiment\_Extractor}}$}, prompted to analyze unstructured news text for tone and key themes. The network topology evolves from a single node to a parallel two-agent structure feeding into the aggregator.

In the subsequent forward pass at iteration $t=1$, the query is routed to both new agents. $A_{\text{Financial\_Analyzer}}$ successfully uses a document parsing tool to extract the financial data. Concurrently, $A_{\text{Sentiment\_Extractor}}$ employs a web search tool to gather and analyze news articles. The aggregator then receives two distinct, high-quality streams of information—quantitative data and qualitative sentiment—which it synthesizes into a comprehensive, nuanced summary that successfully fulfills the user's request. This case demonstrates HiVA's ability to autonomously evolve from a simple to a complex, specialized topology to meet task demands.

\subsection{Case Study 2: Failure Analysis in a Local Optimum}
To understand HiVA's limitations, we examined its performance on a challenging software engineering task: \textit{``Refactor the legacy codebase in `module.py' to use asynchronous calls, while ensuring full backward compatibility with the existing synchronous API."}

HiVA initially demonstrates promise by correctly decomposing the task. It spawns two agents: a \textbf{$A_{\text{Code\_Refactorer}}$} to convert synchronous functions to use `async/await', and a \textbf{$A_{\text{Test\_Writer}}$} to validate the changes. The refactoring agent successfully modifies the code, and the testing agent generates new unit tests that confirm the asynchronous functions work as expected. However, the critical constraint of ``backward compatibility" is not adequately tested by the newly generated tests.

The system receives negative feedback only when the code is evaluated against an external integration test suite, which calls the original synchronous API endpoints. The resulting textual gradient is concise but ambiguous: ` ``Integration tests failed: legacy API call returned an error."' This localized feedback creates a performance trap. The system enters a loop, interpreting the gradient as a flaw in either the refactored code's logic or the new unit tests. The agents make minor, iterative changes—the refactorer might tweak an `await' call, or the tester might adjust an `assert' statement—but neither possesses the global context to diagnose the true architectural issue: the breaking of the synchronous API contract. The system becomes stuck in a local optimum, refining a solution that is internally consistent but externally invalid, highlighting a limitation where localized gradients can fail to capture holistic system constraints.

\subsection{Illustrative Reasoning Trace}
The final, successful reasoning trace for the financial analysis task in our first case study exemplifies a clean and effective multi-agent workflow. The process began with the initial query being routed to two specialized agents simultaneously. The first, \textbf{Agent $A_{\text{Financial\_Analyzer}}$}, received the instruction ``Extract key financial metrics from the provided Q2 2025 earnings report" and utilized its `Document\_Parser' tool, yielding a structured JSON object containing revenue, net profit, and EPS data. The second, \textbf{Agent $A_{\text{Sentiment\_Extractor}}$}, was concurrently instructed to ``Analyze market sentiment for TechCorp in Q2 2025 from financial news" and used its `Web\_Search' tool to produce a summarized list of positive and negative themes. Both of these structured outputs were then passed to the \textbf{Aggregator} agent, which received the final instruction to ``Synthesize the financial data and market sentiment into a summary of risks and opportunities," producing the final, comprehensive analysis.

\section{Ablation Study Details}
This section provides a detailed breakdown of the ablation study presented in the main text (Table~\ref{tab:ablation_study}). We elaborate on the specific configuration of each ablation, offer a detailed analysis of the per-task results, and report on the statistical significance of our findings to elucidate the contribution of each core component of the HiVA framework.

\subsection{Ablation Configurations}
To isolate the impact of each component, we defined five distinct ablation configurations by deactivating a single mechanism from the full HiVA framework:

\begin{itemize}
    \item \textbf{HiVA w/o TEV (Topological Evolution):} In this setting, the agent graph's ability to evolve structurally is disabled. While the system can still refine agent semantics, it cannot create new connections, such as parallel branches for task decomposition. The topology remains fixed to the structure it possessed at the end of the initialization phase (typically a simple linear chain), thus preventing dynamic, self-organizing structural adaptation.

    \item \textbf{HiVA w/o SEV (Semantic Evolution):} Here, the feedback loop for updating agent parameters is severed. The textual gradients generated during the backward pass are ignored, and agents operate with their initial, static system prompts and tool definitions throughout the evaluation. This configuration tests the importance of adaptive agent behavior.

    \item \textbf{HiVA w/o KABB (Knowledge-Aware Routing):} In this configuration, the entire knowledge-aware MAB routing mechanism is disabled. Instead of selectively routing the task to a few relevant agents, the system activates all available successor agents at each step. This transforms the dynamic, sparse execution graph into a dense, broadcast-style workflow, similar to a naive ensemble method. This ablation is designed to measure the efficiency and performance gains derived from intelligent, knowledge-based agent selection compared to a brute-force, non-selective approach.

    \item \textbf{HiVA w/o Env (Environment Feedback):} In this setup, the system is disconnected from the external, ground-truth environment. Instead of receiving objective feedback, the framework falls back to a \textbf{self-evaluation} mechanism to generate textual gradients. After producing a final answer, the aggregator agent critiques its own output based on internal criteria such as logical consistency, completeness, and alignment with the initial instruction. This self-generated critique then serves as the basis for the backward pass. This ablation is designed to measure the performance gap between evolving with objective, external feedback versus a purely subjective, self-correction loop.

    \item \textbf{HiVA w/o Tool (Tool Integration):} In this final ablation, agents are stripped of their ability to generate or execute external tools. They are forced to solve all problems relying exclusively on the intrinsic knowledge and reasoning capabilities of the backbone LLM.
\end{itemize}

\subsection{Detailed Per-Task Analysis}
The quantitative results in Table~\ref{tab:ablation_study} reveal how the importance of each component varies with the task domain.

On \textbf{HotpotQA}, a multi-hop reasoning task, the removal of Topological Evolution (TEV) and Semantic Evolution (SEV) causes the most significant performance drops (-7.3\% and -10.7\%, respectively). This is expected, as successfully answering these questions requires dynamically forming reasoning chains (a topological challenge) and refining agents to be specialists in information extraction and synthesis (a semantic challenge).

On \textbf{MMLU}, which tests broad, multi-domain knowledge, SEV and KABB are shown to be critical. The large performance drop from removing SEV (-5.2\%) indicates that generic agents are insufficient; they must semantically specialize to handle expert-level questions. The degradation from removing KABB (-4.4\% avg.) highlights the necessity of an intelligent routing mechanism to select the correct specialized agent from a diverse pool.

The results on \textbf{MBPP}, a programmatic benchmark, are particularly insightful. While removing semantic and topological evolution still degrades performance, we observe that ablating the tool system results in a slight performance *increase* (+2.2\%). This suggests that for the well-defined coding tasks in this benchmark, the backbone LLM's direct code generation capabilities are highly effective, and the overhead or potential for error in the dynamic tool generation and execution process can occasionally outweigh its benefits.

\subsection{Statistical Significance}
To validate our findings, we conducted paired t-tests comparing the performance of the full HiVA model against each of the five ablated versions on the sampled test sets for HotpotQA, MBPP, and MMLU. The performance degradation for the most critical components, TEV and SEV, was found to be statistically significant across all three benchmarks ($p < 0.01$). The removal of KABB and environment feedback also resulted in statistically significant performance drops in most cases ($p < 0.05$). The slight performance increase observed for the ``HiVA w/o Tool" configuration on MBPP was not statistically significant ($p = 0.12$), suggesting it may be attributable to task-specific variance rather than a systematic advantage.

\section{Algorithmic Details}

This section provides a more detailed description of HiVA's core algorithmic components, focusing on the \texttt{RepairTopology} function as referenced in the main text, and key implementation notes that ensure the framework's robustness and efficiency.

\subsection{The RepairTopology Function}
The \texttt{RepairTopology} function is a crucial maintenance subroutine, invoked specifically after the topology has been modified by the evolution function $f_G$ within an optimization iteration (see Algorithm \ref{algo:stev}). Its primary role is to ensure the agent graph $\mathcal{G}$ remains a valid, efficient Directed Acyclic Graph (DAG). This process involves three main, sequential steps. First, for \textbf{Cycle Prevention}, the function performs a cycle detection check using a standard Depth-First Search (DFS) algorithm. If a back edge is detected—indicating the latest modification created a cycle—that modification is immediately reverted to preserve the DAG property, ensuring predictable information flow. Second, for \textbf{Pruning of Isolated Nodes}, the function identifies any agent $v_i$ (that is not the global source $v_s$ or sink $v_a$) with an in-degree or out-degree of zero. Such nodes are pruned, as they have become disconnected from the workflow and cannot contribute. The removal of a node might break a chain of dependencies; the framework relies on the KABB routing mechanism in the subsequent forward pass to dynamically discover new, valid pathways rather than attempting complex ``path-stitching". Finally, the function performs \textbf{Efficiency-Based Edge Pruning}. It analyzes the historical success rate of each edge, a value tracked across iterations. If an edge's success rate falls below a predefined, tunable hyperparameter (\textit{e.g.}, 0.3), it is pruned. This threshold represents a trade-off between exploration and exploitation; a higher value leads to more aggressive pruning of underperforming pathways. These automated repair mechanisms ensure the agent topology evolves towards efficient and structurally sound configurations.

\subsection{Implementation Notes}
The practical implementation of HiVA relies on several key design choices to manage complexity, ensure stability, and promote scalability. The framework is built with a \textbf{Modular and Asynchronous Architecture}, where each agent is an independent service. This design is realized using Python's \texttt{asyncio} library, enabling non-blocking I/O and parallel execution of agents when the graph topology permits. To mitigate security risks from dynamically generated code, all tool executions are performed within a secure \textbf{Sandboxed Tool Execution} environment, typically implemented using \texttt{Docker} containers or a restricted Python interpreter. This prevents unauthorized file system or network access. A critical component is the \textbf{State Management} system. The entire state of the multi-agent system—including the graph structure $\mathcal{G}$, the Bayesian belief parameters ($\alpha, \beta$), and individual agent prompts—is persisted in a centralized key-value store (like Redis). This allows for stateless agent services and ensures consistency across asynchronous operations. The system also incorporates \textbf{Fault Tolerance} mechanisms; agent or tool invocations have built-in timeout and retry logic (\textit{e.g.}, 3 retries with exponential backoff) to handle transient failures, enhancing overall robustness. Regarding \textbf{Complexity and Scalability}, the worst-case complexity for one iteration is $O(|\mathcal{V}|^2)$. However, KABB routing and aggressive pruning ensure the graph remains sparse in practice. Memory usage scales primarily with the number of agents, peaking at approximately 20GB per GPU on our test hardware for graphs with around 100 agents.

\section{Visualizations}

To complement the formal descriptions provided in the main paper, this section offers visual illustrations of HiVA's core dynamic mechanisms. These diagrams share a unified visual style to provide an intuitive and consistent understanding of how the framework adapts its structure and processes information.

\tikzset{
    base/.style           = {font=\sffamily\small, thick},
    base-arrow/.style     = {-{Latex[length=2.5mm, width=2mm]}, base},
    agent/.style          = {base, rectangle, rounded corners, minimum height=1.1cm, minimum width=2.4cm, align=center, draw},
    agent-inactive/.style = {agent, draw=neutralColor!70, text=neutralColor!80},
    agent-active/.style   = {agent, draw=primaryColor, fill=primaryColor!10},
    arrow-forward/.style  = {base-arrow, draw=primaryColor},
    arrow-feedback/.style = {base-arrow, draw=secondaryColor},
    arrow-neutral/.style  = {base-arrow, draw=neutralColor!80},
    anno-text/.style      = {font=\sffamily\scriptsize, text=neutralColor, align=center},
    feedback-text/.style  = {font=\sffamily\tiny, text=secondaryColor, align=left, inner sep=2pt, midway},
    title-text/.style     = {font=\sffamily\bfseries, text=black!70}
}

\subsection{Topology Evolution}
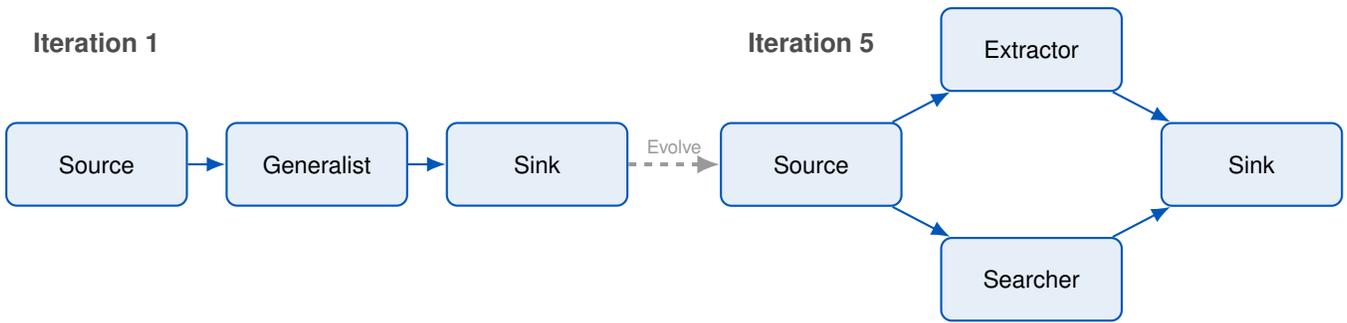
\begin{figure*}[!ht]
\centering
\begin{tikzpicture}[node distance=0.8cm and 0.5cm]
    \begin{scope}[local bounding box=stage1]
        \node[title-text] (t1) {Iteration 1};
        \node[agent-active, below=of t1] (s1) {Source};
        \node[agent-active, right=of s1] (a1) {Generalist};
        \node[agent-active, right=of a1] (k1) {Sink};
        \draw[arrow-forward] (s1) -- (a1);
        \draw[arrow-forward] (a1) -- (k1);
    \end{scope}

    \begin{scope}[xshift=9.5cm, local bounding box=stage2]
        \node[title-text] (t5) {Iteration 5};
        \node[agent-active, below=of t5] (s5) {Source};
        
        \node[agent-active, above right=0.4cm and 0.5cm of s5] (a5u) {Extractor};
        \node[agent-active, below right=0.4cm and 0.5cm of s5] (a5d) {Searcher};
        
        \node[agent-active, right=0.5cm of $(a5u.east)!0.5!(a5d.east)$] (k5) {Sink};
        
        \draw[arrow-forward] (s5) -- (a5u);
        \draw[arrow-forward] (s5) -- (a5d);
        \draw[arrow-forward] (a5u) -- (k5);
        \draw[arrow-forward] (a5d) -- (k5);
    \end{scope}

    \draw[-{Latex[length=3mm, width=2.5mm]}, line width=1.5pt, neutralColor, dashed] 
        ([xshift=0cm]k1.east) -- ([xshift=-0cm]s5.west)
        node[midway, above, anno-text] {Evolve};
\end{tikzpicture}
\caption{An illustrative example of topology evolution. The agent graph transforms from a simple linear chain (left) into a parallelized structure with specialized agents (right), adapting to the demands of the task environment.}
\label{fig:topology_evolution}
\end{figure*}

The agent graph in HiVA is not static; it self-organizes over time to match the complexity of the tasks it encounters. Figure \ref{fig:topology_evolution} illustrates this adaptive process. Initially, the system might use a linear chain with a single `Generalist' agent. As it receives feedback on more complex, decomposable tasks, the topological evolution function ($f_G$) may spawn new, specialized agents in parallel, such as an `Extractor' and a `Searcher'. Over many iterations, the system converges to a complex and efficient topology tailored to the problem domain, demonstrating structural learning.

\subsection{Textual Gradient Flow}
\begin{figure*}[!ht]
\centering
\begin{tikzpicture}[node distance=0.5cm and 2.5cm]
    \node[agent-active] (A1) {Planner};
    \node[agent-active, right=of A1] (A2) {Coder};
    \node[agent-active, right=of A2] (A3) {Aggregator};
    \node[agent-active, right=of A3] (Env) {Environment};

    \draw[arrow-forward] (A1) -- node[midway, above, font=\tiny] {Plan} (A2);
    \draw[arrow-forward] (A2) -- node[midway, above, font=\tiny] {Code} (A3);
    \draw[arrow-forward] (A3) -- node[midway, above, font=\tiny] {Result} (Env.west);

    \draw[arrow-feedback, line width=1.5pt, bend right=25] (Env.north) to node[feedback-text, pos=0.5, below, align=center] {\textbf{Global Feedback:}\\ ``Test Failed"} (A3.north);
    \draw[arrow-feedback, line width=1.5pt, bend right=25] (A3.north) to node[feedback-text, pos=0.5, below, align=center] {\textbf{Gradient for Coder:}\\ ``Logic Error in function X"} (A2.north);
    \draw[arrow-feedback, line width=1.5pt, bend right=25] (A2.north) to node[feedback-text, pos=0.5, below, align=center] {\textbf{Gradient for Planner:}\\ ``Misunderstood Requirements"} (A1.north);
\end{tikzpicture}
\caption{Visualization of the textual gradient backpropagation. A high-level environmental feedback signal is progressively decomposed into specific, actionable critiques for each agent along the reverse execution path, enabling targeted semantic updates.}
\label{fig:gradient_flow}
\end{figure*}
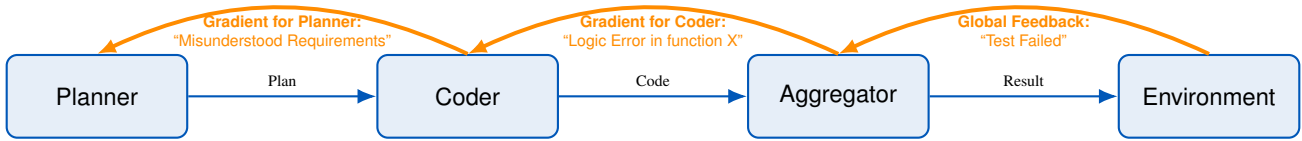

The learning and adaptation in HiVA are driven by textual gradients, which propagate feedback backward through the agent graph. Figure \ref{fig:gradient_flow} visualizes this process. When the final output receives negative feedback from the environment, a high-level error signal travels backward along the execution path. At each step, an LLM decomposes the feedback, creating a more localized and specific gradient for the predecessor agent. This demonstrates how high-level outcomes are translated into actionable, localized feedback for semantic evolution.

\subsection{Dynamic Subgraph Construction}
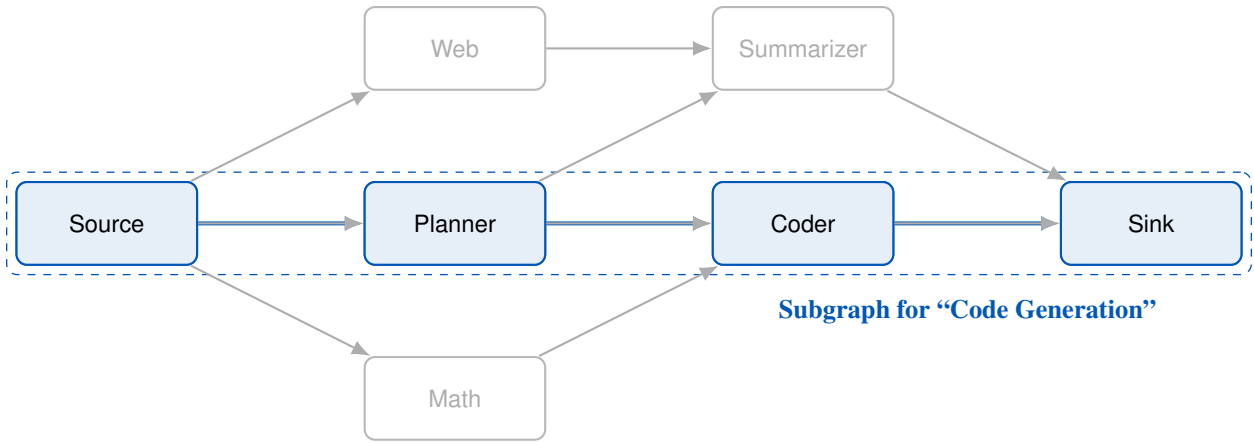
\begin{figure*}[!ht]
\centering
\begin{tikzpicture}[node distance=1.2cm and 2.2cm]
    \node[agent-active] (s) {Source};
    \node[agent-inactive] (a1) [above right=of s] {Web};
    \node[agent-active] (a2) [right=of s] {Planner};
    \node[agent-inactive] (a3) [below right=of s] {Math};
    \node[agent-inactive] (b1) [right=of a1] {Summarizer};
    \node[agent-active] (b2) [right=of a2] {Coder};
    \node[agent-active] (k) [right=of b2] {Sink};

    \draw[arrow-neutral] (s) -- (a1); \draw[arrow-neutral] (s) -- (a2); \draw[arrow-neutral] (s) -- (a3);
    \draw[arrow-neutral] (a1) -- (b1); \draw[arrow-neutral] (a2) -- (b1); \draw[arrow-neutral] (a2) -- (b2);
    \draw[arrow-neutral] (a3) -- (b2); \draw[arrow-neutral] (b1) -- (k); \draw[arrow-neutral] (b2) -- (k);

    \begin{scope}[on background layer]
        \node[agent-active] (s_a) at (s) {Source};
        \node[agent-active] (a2_a) at (a2) {Planner};
        \node[agent-active] (b2_a) at (b2) {Coder};
        \node[agent-active] (k_a) at (k) {Sink};
        \draw[arrow-forward, line width=1.5pt] (s_a) -- (a2_a);
        \draw[arrow-forward, line width=1.5pt] (a2_a) -- (b2_a);
        \draw[arrow-forward, line width=1.5pt] (b2_a) -- (k_a);
        \node[draw, rounded corners, primaryColor, dashed, line width=0.5pt,
              fit=(s_a) (a2_a) (b2_a) (k_a),
              label={[primaryColor, xshift=4.5cm, yshift=-0.2cm]below:\textbf{Subgraph for ``Code Generation"}}] {};
    \end{scope}
\end{tikzpicture}
\caption{Illustration of dynamic subgraph construction. From the complete agent graph, the KABB routing mechanism activates only a relevant subset of agents (highlighted in gray) to handle a specific ``Code Generation" task, thereby creating an efficient and specialized workflow.}
\label{fig:subgraph_construction}
\end{figure*}

For any given task, HiVA does not activate its entire network. Instead, the KABB routing mechanism dynamically constructs a sparse, task-specific execution subgraph. Figure \ref{fig:subgraph_construction} illustrates this. From the complete agent graph (grayed out), the router selects a sequence of the most relevant agents based on the task's needs (\textit{e.g.}, for a ``Code Generation" task). This forms an efficient, temporary subgraph (highlighted in blue) that is used for the forward pass, ensuring both efficiency and accuracy by directing the task to the most suitable specialists.

\section{Baseline Implementation Details}

This section provides detailed descriptions of the configurations, hyperparameters, and models used for the baseline methods, ensuring a fair comparison and reproducibility. All baseline experiments were conducted on the same hardware infrastructure as HiVA, utilizing a cluster of 8 NVIDIA A100 GPUs and Together.AI API.

\subsection{Baseline Configurations}

We implemented a range of baselines, from single-agent prompting strategies to complex multi-agent frameworks. For baselines with mature and publicly available codebases, we utilized their official implementations. In cases where official code was unavailable or could not be reliably run in our environment, we carefully re-implemented the methods, faithfully adhering to the architectures, algorithms, and experimental details described in their respective publications to ensure a high-fidelity comparison.

The baseline setups are as follows. \textbf{Vanilla} represents the direct, zero-shot performance of the backbone LLM, where the task instruction was directly fed to the model. For single-agent methods, \textbf{CoT (Chain-of-Thought)} was implemented by appending ``Let's think step by step." to the prompt. \textbf{Self-Consistency} involved generating five independent reasoning paths ($N=5$) and selecting the final answer by a majority vote. \textbf{Self-Refine} generated an initial output, which was then iteratively improved by the same LLM for a maximum of three iterations ($I=3$). \textbf{TextGrad} was configured as a single-agent optimization method that uses textual gradients to refine its prompt, but lacks topological evolution.

For multi-agent frameworks, \textbf{Multi-Agent Debate} was configured with three agents ($M=3$) engaging in three rounds of debate ($R=3$). For structured frameworks like \textbf{DyLAN}, \textbf{AgentVerse}, \textbf{ADAS}, and \textbf{MaAS}, we followed their described methodologies, utilizing their predefined agent roles and communication protocols. Lastly, \textbf{AutoGPT} was evaluated on the GAIA benchmark using its standard reactive agent loop equipped with fundamental tools like a web searcher and a file system interface.

\subsection{Hyperparameter Settings}

To ensure a fair comparison, we standardized key hyperparameters across all baselines wherever possible, consistent with the settings used for HiVA. For all LLM inference, the decoding temperature was set to \textbf{1.0} to encourage output diversity, and the maximum number of generated tokens was capped at \textbf{4096} to accommodate complex tasks. Method-specific parameters included using five reasoning paths ($N=5$) for Self-Consistency, three refinement iterations ($I=3$) for Self-Refine, and a configuration of three agents engaging in three debate rounds ($R=3$) for Multi-Agent Debate. For other complex frameworks, we adopted the default hyperparameter settings recommended in their documentation or publications to represent them under their optimal conditions.

\subsection{Model Versions and Infrastructure}

The primary large language model used for all experiments was \texttt{Qwen-2.5-72B-Instruct-Turbo}. For specific agentic tasks as noted in the main text, \texttt{GPT-4o-mini} was also employed. These models served as the backbone for all baseline methods and the agents within HiVA.

\section{Mathematical Derivations}

This section provides a formal exposition of the key mathematical and algorithmic constructs that underpin the HiVA framework, elaborating on the concepts introduced in the main text.

\subsection{Generalized Gradient Descent in Hybrid Space}
The core optimization problem in HiVA is to find an optimal system configuration $s^*$ that minimizes a black-box objective function $\mathcal{L}(s)$, which measures performance in a given environment. The solution $s$ exists in a hybrid space $\mathcal{S} = \mathcal{G} \times \mathcal{P}_\Theta$, where $\mathcal{G}$ represents the discrete space of possible agent graph topologies and $\mathcal{P}_\Theta$ is the semantic space of agent parameters (\textit{e.g.}, prompts, tool configurations).
The primary challenge is that $\mathcal{S}$ is non-Euclidean, discrete, and non-differentiable, rendering traditional gradient descent inapplicable. To overcome this, we formulate the optimization as a \textbf{Generalized Gradient Descent} process. Instead of a numerical gradient vector, we introduce the concept of a \textbf{Textual Gradient}, $\Delta s_t$, which is a structured command object generated by a Large Language Model acting as a \textbf{Textual Gradient Parser} (TGP).
The update rule is defined as a symbolic operation:
$$
s_{t+1} \leftarrow s_t \oplus \Delta s_t
$$
Here, $\oplus$ denotes the application of the structured commands in $\Delta s_t$ to the current configuration $s_t$. The textual gradient is a composite object $\Delta s_t = (\Delta \mathcal{G}_t, \Delta \mathcal{P}_{\Theta,t})$, where $\Delta \mathcal{G}_t$ contains topological commands (\textit{e.g.}, `ADD\_SUCCESSOR', `REMOVE\_EDGE') and $\Delta \mathcal{P}_{\Theta,t}$ contains semantic modification instructions (\textit{e.g.}, prompt rewrite directives).

This process is guided by a ``textual chain rule" during the backward pass. First, a global textual gradient $\nabla_{\text{text}}\mathcal{L}_t$ is generated at the aggregator agent $v_a$ based on environmental feedback. This gradient is then propagated backward through the execution graph. For any agent $v_i$, its localized gradient $\frac{\partial\mathcal{L}_t}{\partial v_i}$ is derived by an LLM-based function that synthesizes the feedback from its successors and its own output:
$$
\frac{\partial\mathcal{L}_t}{\partial v_i} \approx \text{LLM}\left(\left\{\frac{\partial\mathcal{L}_t}{\partial v_j} \mid v_j \in \text{successors}(v_i)\right\}, y_i\right)
$$
This localized gradient then informs the generation of the specific update commands in $\Delta s_t$ for agent $v_i$.

\subsection{Bayesian Update for Knowledge-Aware Routing}
The agent selection process in the forward pass is managed by the Knowledge-Aware Bayesian-Bandit (KABB) routing mechanism, which models the problem as a Multi-Armed Bandit solved with Thompson Sampling. At each step, the probability of selecting an agent $A_i$ is proportional to a score that balances historical performance, task relevance, and team synergy.
The update rules for the Bayesian belief parameters, $\alpha_i$ and $\beta_i$ (representing success and failure counts for agent $A_i$), are central to this process. These parameters evolve according to the following equations, which incorporate a reward signal, a knowledge-driven adjustment, and a temporal decay factor:
$$
\alpha_i^{(t+1)} = \gamma^{\Delta t} \alpha_i^{(t)} + \left[ r_i^{(t)} + \delta \cdot \text{KM}(A_i, I_{\text{task}}) \right] \cdot \mathbb{I}_{\{A_i \in \mathcal{S}_t\}}
$$
$$
\beta_i^{(t+1)} = \gamma^{\Delta t} \beta_i^{(t)} + \left[ 1 - r_i^{(t)} + \delta \cdot \text{KD}(A_i, I_{\text{task}}) \right] \cdot \mathbb{I}_{\{A_i \in \mathcal{S}_t\}}
$$
In these equations, $r_i^{(t)} \in \{0, 1\}$ is the observed reward for agent $A_i$'s contribution in iteration $t$. The term $\gamma^{\Delta t} = e^{-\kappa \Delta t}$ is an exponential decay factor that prioritizes recent information, where $\kappa$ is a decay constant and $\Delta t$ is the time elapsed. The indicator function $\mathbb{I}_{\{A_i \in \mathcal{S}_t\}}$ ensures that only agents selected for the current task are updated. The terms $\text{KM}(A_i, I_{\text{task}})$ and $\text{KD}(A_i, I_{\text{task}})$ represent a knowledge-driven bonus and penalty, respectively, which are derived from an external knowledge graph to measure the semantic alignment between an agent's capabilities and the task's requirements. The hyperparameter $\delta$ controls the influence of this knowledge-based component. This formulation allows the system to make informed decisions even for novel tasks where historical performance data ($r_i^{(t)}$) is sparse.

\subsection{Evolution of Structural Memory Weights}
The agent graph itself functions as a form of distributed memory, where the strength of inter-agent collaborations is encoded in edge weights. As described in the main text, the synergy coefficient $C_{\text{syn}}(v_i, v_j)$ for an edge from agent $v_i$ to $v_j$ evolves over time. This update rule combines the existing weight with a term representing the most recent performance of that specific interaction:
$$
C_{\text{syn}}^{(t+1)}(v_i, v_j) = C_{\text{syn}}^{(t)}(v_i, v_j) + \gamma' \cdot \frac{\alpha_{ij}^{(t)}}{\alpha_{ij}^{(t)} + \beta_{ij}^{(t)}} \cdot \mathcal{R}_{ij}^{(t)}
$$
Here, $\gamma'$ is a learning rate for the synergy update. The term $\frac{\alpha_{ij}^{(t)}}{\alpha_{ij}^{(t)} + \beta_{ij}^{(t)}}$ is the expected success rate of the specific interaction path $(v_i, v_j)$, estimated from a dedicated Beta distribution with parameters $\alpha_{ij}$ and $\beta_{ij}$ that track that edge's historical performance. $\mathcal{R}_{ij}^{(t)}$ is a measure of the utility or contribution of the edge in the $t$-th iteration, often derived from the textual gradient analysis. This mechanism ensures that frequently used and successful collaboration pathways are structurally reinforced over time.

\end{document}